\documentclass[10pt,twocolumn,letterpaper]{article}

\usepackage{cvpr}              % To produce the CAMERA-READY version
\usepackage[dvipsnames]{xcolor}

\usepackage[accsupp]{axessibility}
\definecolor{cvprblue}{rgb}{0.21,0.49,0.74}
\usepackage[pagebackref,breaklinks,colorlinks,citecolor=cvprblue]{hyperref}
\usepackage{amsmath}
\usepackage[utf8]{inputenc} % allow utf-8 input
\usepackage[T1]{fontenc}    % use 8-bit T1 fonts
\usepackage{hyperref}       % hyperlinks
\usepackage{url}            % simple URL typesetting
\usepackage{booktabs}       % professional-quality tables
\usepackage{amsfonts}       % blackboard math symbols
\usepackage{nicefrac}       % compact symbols for 1/2, etc.
\usepackage{microtype}      % microtypography
\usepackage{xcolor}  
\usepackage{colortbl}
\usepackage{enumitem} 
\usepackage{caption}
\usepackage{multirow}
\usepackage{float}
\usepackage{arydshln}
\usepackage{amssymb}
\usepackage{graphicx}
\usepackage{pifont}
\definecolor{mygreen}{rgb}{0.1, 0.7, 0.2}
\definecolor{LightCyan}{rgb}{0.96,0.96,0.96}
\definecolor{lightblue}{RGB}{240,240,240}
\newcommand{\hn}{\textit{hard negatives }}

\def\paperID{7445} % *** Enter the Paper ID here
\def\confName{CVPR}
\def\confYear{2024}

\title{Contrasting intra-modal and ranking cross-modal hard negatives to enhance visio-linguistic compositional understanding}

\author{%
  Le Zhang$^{1,2}$\quad Rabiul Awal$^{1}$ \quad Aishwarya Agrawal$^{1, 2, 3}$ \\
  Mila - Quebec AI Institute$^1$ \\ Université de Montréal$^2$ \\ 
 Canada CIFAR AI Chair$^3$
}

\begin{document}
\maketitle
\begin{abstract}
Vision-Language Models (VLMs), such as CLIP, exhibit strong image-text comprehension abilities, facilitating advances in several downstream tasks such as zero-shot image classification, image-text retrieval, and text-to-image generation. However, the compositional reasoning abilities of existing VLMs remains subpar. The root of this limitation lies in the inadequate alignment between the images and captions in the pretraining datasets. Additionally, the current contrastive learning objective fails to focus on fine-grained grounding components like relations, actions, and attributes, resulting in "bag-of-words" representations. We introduce a simple and effective method to improve compositional reasoning in VLMs. Our method better leverages available datasets by refining and expanding the standard image-text contrastive learning framework. Our approach does not require specific annotations and does not incur extra parameters. When integrated with CLIP, our technique yields notable improvement over state-of-the-art baselines across five vision-language compositional benchmarks. \footnote{We open-source our code at \url{https://github.com/lezhang7/Enhance-FineGrained}.}
\end{abstract}

\section{Introduction}

The field of vision-language research has experienced remarkable progress over recent years, thanks to the introduction of vast datasets \cite{schuhmann2021laion400m,gadre2023datacomp}, the adaptation of attention mechanism, and the pioneering objectives such as contrastive learning. Impressively, these models demonstrate a notable capability in zero-shot generalization, as seen in areas like Visual Question Answering (VQA) \cite{vqa2017}, captioning \cite{blip2,mapl,alayrac2022flamingo}, and image-text retrieval \cite{radford2021learning,zeng2022multigrained,tan2019lxmert}. Strong Vision-Language Models (VLMs), such as CLIP \cite{radford2021learning}, are even pushing the boundaries in text-to-image generation (CLIP is used to guide image generation given the input prompt) \cite{latentdiffusion,dalle,poole2022dreamfusion}.  However, despite these advances, a notable limitation persists: these models often miss the intricate compositional nuances of relationships, attributes, objects, and actions \cite{yuksekgonul2022and,thrush2022winoground}. A clear manifestation of this shortcoming is their difficulty in distinguishing between captions with the same set of words but composed differently like ``Horse is eating the grass'' and ``Grass is eating the horse'' \cite{yuksekgonul2022and} when paired with relevant images. Such compositional understanding remains a critical frontier for continued advancement in vision-language integration.

\begin{figure}[t]
    \centering
    \includegraphics[width=\linewidth]{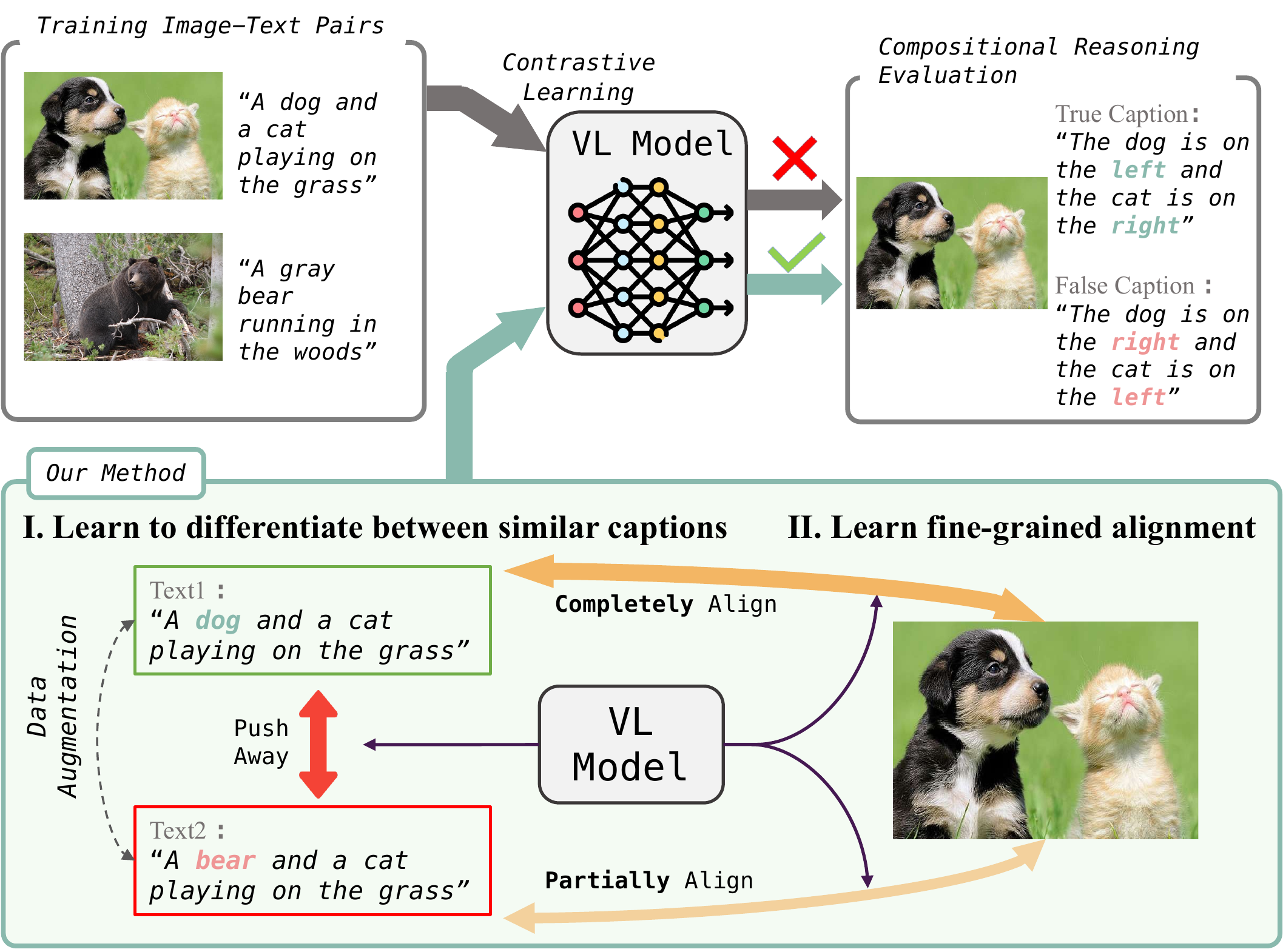}
    \captionsetup{font=small}
    \caption{Models trained with standard image-text contrastive learning lack sufficient compositional reasoning abilities. Our method teaches the model to better differentiate between similar captions and learn fine-grained alignment between images and text to improve compositional reasoning. }
    \label{fig:motivation}
    \vspace{-5mm}
\end{figure}
A primary factor impeding compositional understanding in current VLMs stems from their learning methodology and training dataset. These models are usually trained on huge image-text pairs crawled from the web using contrastive learning \cite{radford2021learning}. The caption is short and noisy; the image-text contrastive objective optimizes the model to distinguish between correct image-text pairs and a vast array of incorrect ones. However, because the incorrect pairs are often markedly distinct, the model primarily distinguishes them through simple object recognition, without needing to comprehend fine-grained details  such as attributes and relations. \cref{fig:motivation} depicts a scenario where CLIP struggles with the compositional reasoning of ``left'' and ``right'' concepts.

% Earlier studies, including NegCLIP \cite{yuksekgonul2022and}, have employed phrase swapping to produce additional captions for training. This underscores the importance of incorporating hard negatives in vision-language contrastive learning. However, when examining the intra-model similarity in models such as CLIP and NegCLIP (refer to \cref{fig:motivation} c), captions with analogous structures (e.g., "Horse is eating the grass" versus "Grass is eating the horse") exhibit closely related representations despite their differing semantics. We propose that a greater disparity in the representations of such analogous captions could enhance the accuracy of text-image alignment, thereby improving vision-language compositional understanding. Furthermore, as depicted in Fig \ref{fig:motivation} (Right), there exists only a slight difference between the similarities of correct image-text pairs and their hard negative counterparts. This observation indicates potential limitations in the model's grasp of compositional alignment, suggesting that hard negatives may not be optimally utilized.

 Earlier studies, including NegCLIP \cite{yuksekgonul2022and}, have employed phrase swapping to produce additional captions for training. This underscores the importance of incorporating hard negatives in vision-language contrastive learning. However, simply incorporating additional samples into standard image-text contrastive learning does not fully leverage hard negatives. In this work, we refine and expand the contrastive learning objective for hard negative captions (see \cref{fig:motivation}), which vary in semantics like relations, attributes, actions, and objects. We focus on two dimensions. First, we advocate for a clearer distinction in the representations of positive and hard-negative captions, aiming to boost the model's ability to recognize nuanced semantic variations. Second, we maintain a minimum similarity gap between authentic image-text pairs and their challenging hard-negative counterparts to encourage the learning of fine-grained image-text alignment. Consequently, we propose two objectives: i) intra-modal contrast, and ii) cross-modal rank, built on the hinge loss~\cite{cortes1995support} approach. The latter incorporates an adaptive threshold during the fine-tuning phase. This means as the model becomes more adept, the threshold increases, reflecting both the growing difficulty of the task and the model's increasing competency. This approach not only resonates with curriculum learning principles but also ensures a more stable training process.

To validate the effectiveness, we conduct experiments on two models: the versatile CLIP and the strong X-VLM \cite{zeng2022multigrained}. Our evaluation across various compositional datasets consistently reveals performance enhancements, establishing our method as a new \textit{state-of-the-art} across all assessed benchmarks. Specifically, training CLIP with our method on the COCO dataset leads to an improvement of 23.7\% and 13.5\% respectively on the Relation and Attribution splits of the ARO benchmark \cite{yuksekgonul2022and}, 7.2\% on the VALSE benchmark~\cite{parcalabescu-etal-2022-valse}, 5.9\% on the VL-CheckList benchmark~\cite{zhao2022vlchecklist}, and a significant improvement of 12.1\% on the recently developed SugarCrepe benchmark \cite{hsieh2023sugarcrepe}. We also achieve modest improvements of 0.5\%, 2.5\% respectively on the ARO Relation and Attribution splits, 1.3\% on VALSE and 2.1\% on VL-CheckList on top of the already strong X-VLM model upon application of our method. Finally we also evaluate our method on the conventional image-text retrieval and image classification benchmarks, resulting in 7.5\% improvement in image-text retrieval and a small 1.6\% decrease in image classification. 

% To summarize, the main contributions of this article are: (1) We devise a method for generating hard negatives that overcomes the existing limitations in ITC loss. This method incorporates a variety of negatives based on distinctive features. (2) We introduce an intra-modal contrastive loss that leverages the generated hard negatives to tackle the challenge of high text similarity in VLMs. (3) We propose a cross-modal rank loss with threshold, encouraging the model to maintain a disparity between cross-modal positive and hard negative similarities, and (4) We make the threshold to be adaptive, acting as a form of curriculum learning that adjusts to the model's evolving capability. (5) We establish enhanced performance on multiple benchmarks using our proposed losses. This assertion is supported by extensive experiments and ablation studies, which illuminate the efficacy and design of our approach.

To summarize, we present three key contributions: (1) We propose a simple yet effective solution to better leverage available image-text datasets to improve VLMs' compositional understanding without introducing any additional parameters. This is achieved by extending the contrastive learning framework: introducing intra-modal contrast and cross-modal rank objectives. (2) Our adaptive threshold strategy induces curriculum learning during fine-tuning, leading to improved results and stable training without the need for labour-some and time-consuming parameter tuning. (3) We demonstrate the effectiveness of our approach through its state-of-the-art performance on five benchmarks. Furthermore, we conduct a thorough analysis of each component of our model, providing insights for future research and a deeper understanding of our methodology through extensive experiments.

\section{Related Work}
\paragraph{Contrastive Vision-Language Models} Vision-language models have garnered remarkable success in both the vision and multimodal domains. Modern VLMs are pretrained on large and noisy multi-modal datasets \cite{schuhmann2022laion5b,schuhmann2021laion400m} and then applied to downstream tasks in a zero-shot manner. Among them, CLIP \cite{radford2021learning} stands out, employing a contrastive learning method for pretraining. Our reasons to focus on CLIP are twofold: firstly, image-text contrastive learning has become a prevalent strategy for VLM pretraining \cite{align,zeng2022multigrained,coca,singh2022flava,zhai2023sigmoid,sun2023evaclip}; secondly, CLIP boasts extensive applicability, spanning various domains. This includes zero-shot image classification \cite{coop, clip-adapter,Metzen2023AutoCLIPAZ,Novack2023CHiLSZI}, object detection \cite{OWL-ViT}, semantic segmentation \cite{maskclip,zhou2023zegclip,Xu2021ASB,wang2023samclip}, text-image retrieval, evaluation of text-image alignment \cite{hessel2022clipscore,Cho2022FinegrainedIC}, and text-to-image generation \cite{latentdiffusion,dalle,poole2022dreamfusion}. Furthermore, the vision encoder from CLIP can serve as a strong backbone for generative vision-language models \cite{blip2,awadalla2023openflamingo,zhu2023minigpt4,llava}. Therefore, enhancements on CLIP can effectively radiate to a broader range of vision-language applications.

% VLMs such as CLIP \cite{radford2021learning}, ALIGN \cite{align}, X-VLM \cite{zeng2022multigrained}, COCA\cite{coca}, FLAVA\cite{singh2022flava} typically encompass a text encoder and an image encoder. These models are trained on extensive, noise-rich multi-modal datasets utilizing a combination of loss functions like Grounded (Masked) Language Modeling, Masked Image Modeling, Image Text Matching, and Image Text Contrastive Learning. The Image Text Contrastive (ITC) loss \cite{radford2021learning} is the most common, as it strives to minimize the distance between aligned image-text pairs and maximize the distance between unaligned pairs. These unaligned pairs, also known as negatives, are randomly sampled within the batch. Due to the large-scale pretraining datasets, VLMs have demonstrated exceptional zero-shot performance across diverse visual-language tasks \cite{radford2021learning,align,singh2022flava}. However, recent research indicates that these models often learn a "bag of words" representation \cite{doveh2022teaching, yuksekgonul2022and}, limiting their compositional understanding.
 \begin{figure*}[h]
    \centering
    \includegraphics[width=0.75\linewidth]{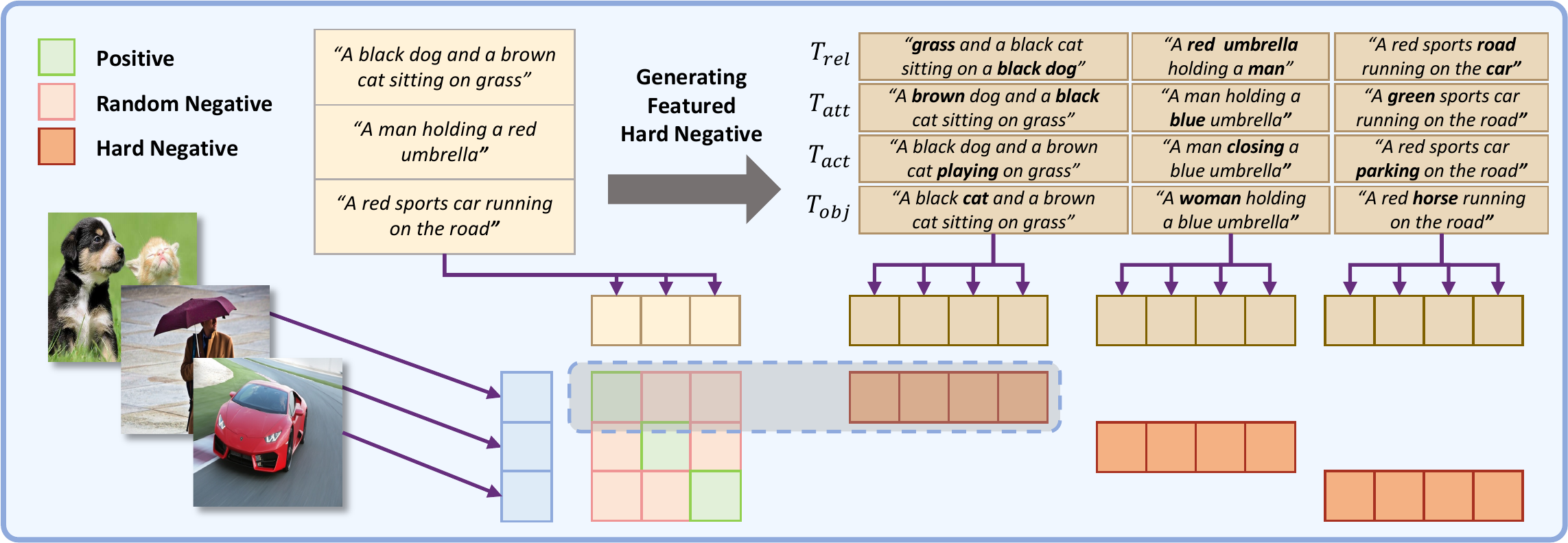}
    \captionsetup{font=small}
    \caption{(Top) An overview of our method's pipeline and hard negative generation examples. Losses are applied on the shaded boxes.}
    \label{fig:hardnegative}
    \vspace{-5mm}
\end{figure*}
    
\paragraph{Vision-Language Compositionality}

   While Vision-Language Models exhibit remarkable strength in handling multimodal data, recent investigations suggest that these models tend to learn a ``bag of words'' representation, which hampers their compositional understanding \cite{doveh2022teaching, yuksekgonul2022and}. A number of benchmarks have emerged to evaluate the performance of VLMs, focusing on various dimensions like relations, attributes, objects, among others. For instance, ARO \cite{yuksekgonul2022and} emphasizes the understanding of attributes and relations, while VL-checklist \cite{zhao2022vlchecklist} drills down into finer subcategories such as size, color, action, and spatial relations. VALSE \cite{parcalabescu-etal-2022-valse} targets linguistic phenomena like existence, counting, plurality, and coreference, whereas Winoground \cite{thrush2022winoground} delves into complex reasoning, encompassing commonsense and external knowledge. SugarCrepe \cite{hsieh2023sugarcrepe} aims to address the hackability issue where pure-text models without image information can outshine robust VLMs on several compositional benchmarks, attributing to a significant distribution gap between positive and hard negative captions. All these benchmarks are structured as cross-modal retrieval tasks -- discern between correct and incorrect captions given an image, and evaluations are based on accuracy metrics.

The quest to augment VLMs' compositional understanding has ignited substantial interest within the community. The DAC approach \cite{DAC} proposes to enhance caption density and quality by utilizing an off-the-shelf caption model \cite{blip2} and a segmentation model \cite{sam}. Conversely, SGVL \cite{SGVL} and MosaiCLIP \cite{Singh2023CoarsetoFineCL} employ additional scene graph annotations to guide model learning on compositional relations. Although these methods demonstrate effectiveness, they necessitate either a specific model (like a segmentation model) or additional annotations (such as a scene graph). A distinct line of research explores hard negative mining methodology \cite{kalantidis2020hard}, where SLVC \cite{doveh2022teaching}, \citet{paiss2023teaching} and NegCLIP \cite{yuksekgonul2022and} enrich samples with negative text via random word-swapping. We perceive negative augmentation as a refined method since it does not hinge on extra resources (model or data) and postulate that the current methodologies do not entirely harness the potential of hard negative mining, and thus, we introduce two additional losses atop our featured hard negatives to further bolster the compositional understanding capability.

\section{Method}

\begin{figure*}
    \centering
    \includegraphics[width=0.7\linewidth]{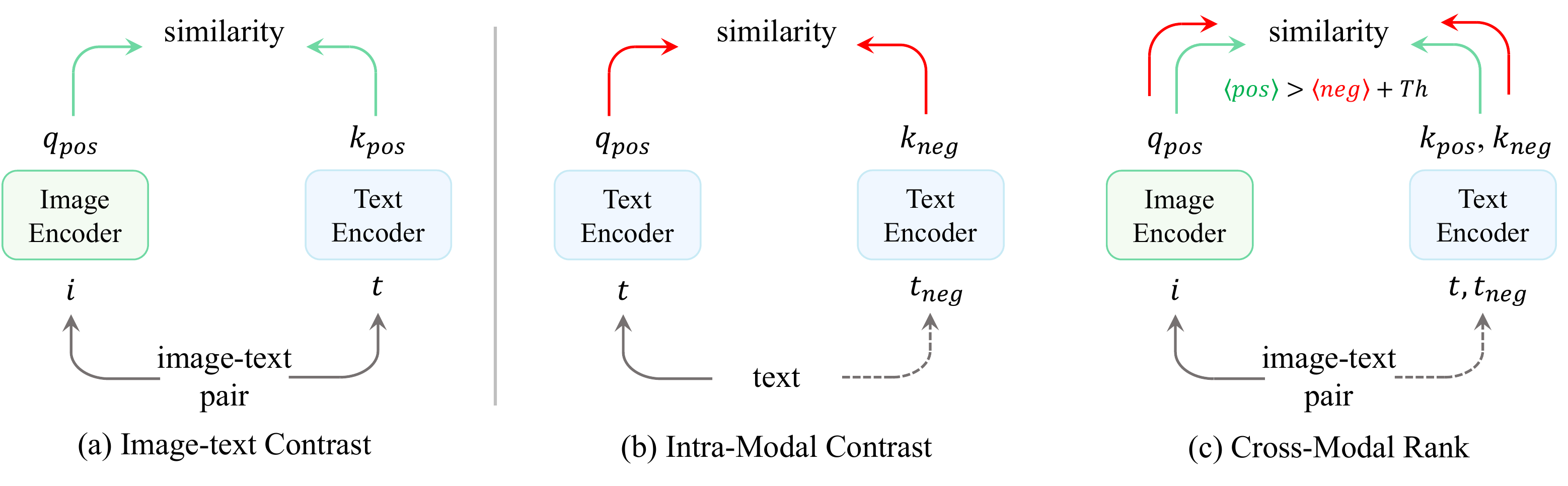}
    \captionsetup{font=small}
    \caption{\textbf{Conceptual loss comparison}. Red arrows denote minimizing similarity, while green arrows denote maximize it; Dotted arrow represents data augmentation. (a) Standard image-text contrastive learning applied in \cite{radford2021learning}. (b) Proposed intra-modal contrast applied on generated hard negative texts and (c) cross-modal rank applied on positive and hard negative pairs with adaptive threshold.}
  
    \label{fig:loss}
    \vspace{-3mm}
\end{figure*}

In the proposed method, we expand upon image-text contrastive learning and introduce two loss functions specifically applied to the automatically generated hard negatives. In this section, we first discuss the process of hard negative generation, followed by a detailed description of our loss functions. \cref{fig:hardnegative} illustrates the overview of pipeline and \cref{fig:loss} illustrates proposed losses.

\subsection{Featured Hard Negative Generation}
\label{sec:generation}
In contrastive learning, \hn  refer to instances that exhibit high similarity to positive samples, yet do not qualify as positive themselves. Consider the following caption as an example: ``A gray cat sits on top of a \textbf{wooden} chair near a plant.'' A potential hard negative could be: ``A gray cat sits on top of a \textbf{plastic} chair near a plant.'' \cite{doveh2022teaching} While the hard negative correctly identifies the majority of elements in the image, it diverges from the positive sample with regards to the chair's material. Incorporating hard negatives into the training process can enable models to discern subtle distinctions, thereby enhancing their overall accuracy and performance \cite{Harwood2017SmartMF,Ge2018DeepML,Manmatha2017SamplingMI,Qin2021UnderstandingAI,Kalantidis2020HardNM,Robinson2020ContrastiveLW}.

To bolster the compositional understanding of our model, we deliberately create hard negatives that embody various alterations to the original captions. These adjustments encompass changes in the relationship, attributes, and action of the image's objects. Furthermore, we produce hard negatives where we replace an object name with another, encouraging the model to distinguish between different objects. To generate these hard negatives, we employ Part-Of-Speech (POS) parsing and Language Models. Utilizing Spacy \cite{spacy2}, we parse the captions and assign POS tags to each word. For relational hard negative, we interchange the positions of two noun words. For attribution, action, and object name alterations, we randomly mask an adjective, verb, or noun word, and subsequently fill in the masked area using the RoBERTa \cite{liu2019roberta}, examples are shown in Fig \ref{fig:hardnegative}. For each caption, we generate all four types of hard negatives, replacing any examples in which the requisite words or two objects are absent from the caption with a placeholder string. This approach ensures a comprehensive and robust training dataset for enhancing our model's performance.

\subsection{Expanded Losses}
\paragraph{Preliminaries}
Contrastive VLMs consist of a image encoder $f_i: X_{image} \longrightarrow \mathbb{R}^d$ and a text encoder $f_t: X_{text} \longrightarrow \mathbb{R}^d$. The cosine similarity between two inputs I, T using their encoders $f_i, f_t$ are computed as:
$S(I,T)=\frac{f_i(I)\cdot f_t(T)}{||f_i(I)||\cdot||f_t(T)||}/\tau$
where $\cdot$ represents inner product and $\tau$ is a trainable temperature parameter. The image-text contrastive loss is applied on the computed similarity. Considering image-text pairs $(I,T)$ within a batch $\mathcal{B}$, the computation of the Image-Text Contrastive (ITC) loss is formulated as follows:
{
\small
\begin{equation}
        \mathcal{L}_{itc}=\sum\limits_{(I,T) \in \mathcal{B}}-\left (\log \frac{\exp^{S(I,T)}}{\sum\limits_{{T_{i} \in \mathcal{B}}}\exp^{S(I,T_i)}}+\log \frac{\exp^{S(I,T)}}{\sum\limits_{{I_{j} \in \mathcal{B}}}\exp^{S(I_j, T)}} \right )
\end{equation}
}

For each image-text pair $(I,T)$, prior research methodologies generate a single hard negative caption $T_{hn}$ through the random swapping of a word. This generated caption is subsequently treated as an additional random negative \cite{yuksekgonul2022and,doveh2022teaching,paiss2023teaching}. Thus, the formulation of the Image-Text Contrastive loss with the inclusion of a hard negative can be described as follows:
\vspace{-1mm}
{
\small
\begin{equation}
\begin{split}
&\mathcal{L}_{itc(hn)}=\sum\limits_{(I,T) \in \mathcal{B}}-\log \frac{\exp^{S(I,T)}}{\sum\limits_{{I_{j} \in \mathcal{B}}}\exp^{S(I_j,T)}} \\
&+\sum\limits_{(I,T) \in \mathcal{B}}- \log \frac{\exp^{S(I,T)}}{\sum\limits_{{T_{i} \in \mathcal{B}}}\exp^{S(I,T_i)}+\sum\limits_{T_{k} \in \mathcal{T}_{hn}}\exp^{S(I,T_{k})}} 
\end{split}
\end{equation}
}

\paragraph{Intra-Modal Contrastive}
Adhering to the aforementioned notations and given an image-text pair $(I,T)$ within batch $\mathcal{B}$, our method, as outlined in Section \ref{sec:generation}, generates four distinct hard negatives $\mathcal{T}_{hn}=\{T_{rel}, T_{att}, T_{act}, T_{obj}\}$ corresponding to changes in relation, attribute, action and object entity. The primary motivation behind employing intra-modal contrastive (IMC) loss is to promote the model's ability to differentiate between hard negative captions to the maximum extent and contrastive loss is well-suited for this purpose.  Consequently, the formulation is:

{
\small
\begin{equation}
    \mathcal{L}_{imc}=\sum\limits_{(I,T) \in \mathcal{B}}-\log \frac{1}{\sum\limits_{{T_{k} \in \mathcal{T}_{hn}}}\exp^{S(T,T_{k})}}
    \label{eq:intro-modal}
\end{equation}
}

\paragraph{Cross-Modal Rank with Adaptive Threshold}

Hard negative captions retain some elements of truth about the image, indicating a partial correctness in the image-text alignment. The model is designed to discern the similarity between a true image-text pair and a hard negative pair to a certain extent; i.e. it stops further optimization using hard negative pairs once the similarity difference exceeds a predefined threshold. To achieve this, we employ a ranking loss with a threshold. This threshold ensures that the similarity score for an image-text pair, \( S(I, T) \), is greater than the similarity score for that image and any hard negative caption, \( S(I, T_{k}) \), by at least a threshold value \( Th_k \) corresponding to the type of hard negative. This concept is formally represented as follows:
\[ S(I, T) > \{ S(I, T_{k}) + Th_k | T_{k} \in \mathcal{T}_{hn} \} \]

% Hard negative captions contain elements of truth about the corresponding image, suggesting they are partially correct in terms of image-text alignment. The model should discern the difference of similarity of true and hard negative image-text pair to a specific degree – i.e., when the difference surpasses a predefined degree – it refrains from further optimizing using hard negative pairs. To this end, we implement a rank loss with a tailored threshold approach. The threshold ensures that the similarity score of an image-text pair $S(I, T)$ exceeds the similarity of that image and its hard negative caption $S(I, T_{k})$ by at least a threshold value $Th_k$ corresponding to the type of hard-negative. This can be formally expressed as:
% \[S(I, T)>\{S(I, T_{k})+Th_k|T_{k} \in \mathcal{T}_{hn}\}\]

Inspired by the hinge loss concept \cite{cortes1995support}, we employ this threshold in the loss function, which we call Cross-modal Rank (CMR) loss, defined as follows:

{
\small
\begin{equation}
    \mathcal{L}_{cmr}=\sum\limits_{(I,T) \in \mathcal{B}} \sum\limits_{{T_{k} \in \mathcal{T}_{hn}}}max(0,S(I, T_{k})-S(I, T)+Th_{k})
\end{equation}
}

% Determining an appropriate threshold for hinge loss presents a considerable challenge \cite{wang2014learning}. Existing research on adaptive thresholds \cite{chen2017triplet,zhao2019weakly,Liu_2019_CVPR} has indicated that an effective threshold should evolve in accordance with the training progress. In this context, we adapt this principle to the multi-modal learning domain. Specifically, we model the threshold based on the disparity in the model's similarity scores between the true pair and the \hn pair. Importantly, this disparity can be interpreted as an indicator of the model's capacity for compositional understanding. In the early stages of training, when the model struggles to differentiate between the \hn and true pairs, a lower threshold is appropriate. As the model becomes more adept in its compositional understanding throughout training, it should demonstrate a growing difference in scores between the \hn and true pairs. This optimization strategy aligns with the principles of curriculum learning. By progressively adjusting the threshold, we anticipate enhanced convergence, diminished overfitting, and improved generalization \cite{wang2021survey}.
% 
% In this context, the threshold reflects both the task's complexity and the model's capability. Therefore, at training step $t$, the threshold for a specific type $\{k|k \in (rel,att,act,obj)\}$ is calculated as follows:

Determining an appropriate threshold for hinge loss is challenging \cite{wang2014learning}. Inspired by existing research on adaptive thresholds \cite{chen2017triplet,zhao2019weakly,Liu_2019_CVPR,zhang2018collaborative} that posit that an effective threshold should evolve in accordance with the training progress, we adapt this principle to the multi-modal learning domain. Our approach models the threshold using the difference in the model's similarity scores between the true and \hn pairs, serving as a indicator of the model's compositional understanding capability. Especially during the initial training phase, when differentiating between the \hn and true pairs is tough, a lower threshold is appropriate. As training advances, and the model refines its understanding, this score disparity grows. This progressive threshold adaptation, aligning with curriculum learning principles, aims for smoother optimization, avoidance of local minima, and improved generalization \cite{wang2021survey}. 
Consequently, the threshold encapsulates both task intricacy and model proficiency. Thus, at training step 
$t$, the threshold for each type $\{k|k \in (rel,att,act,obj)\}$ is computed as:
{
\small
\begin{equation}
    Th_k^{t}=\frac{1}{|\mathcal{B}|}\sum_{(I,T) \in \mathcal{B}}(S^{t-1}(I,T)-S^{t-1}(I,T_k))
\end{equation}
} 
Another unique aspect of our approach is that we implement distinct thresholds for different types of hard negatives, each tailored to a specific ``curriculum'', while most existing approaches utilizing adaptive thresholds in non-multimodal domains \cite{chen2017triplet,zhao2019weakly,Liu_2019_CVPR,zhang2018collaborative} employ just one threshold. The adaptive Cross-modal Rank loss at step $t$ is defined as:
{
\small
\begin{equation}
\label{eq:cmr}
    \mathcal{L}_{cmr}=\sum\limits_{(I,T) \in \mathcal{B}} \sum\limits_{{T_{k} \in \mathcal{T}_{hn}}}max(0,S(I, T_{k})-S(I, T)+Th^t_{k})
\end{equation}
}
% We empirically observed that if there is no constraint placed on the threshold, the threshold for relation hard negatives (\hn) tends to escalate too rapidly, thereby disrupting the training process. The reason behind is that unlike other hard neg this is not created by replacing original word with plausible words, so this hard neg may be very implausible and hence easier for the model to distinguish it from true pair leading to high disparity between similarities. To stabilize the training, it's therefore crucial to confine the threshold within an upper limit $u$.

Empirically, we find that adding the term $-S(T,T_{rel})$ to CMR offers benefits and without threshold constraints, the value of relation hard negatives escalates rapidly, hindering training. This is because these negatives, unlike others, are not formed by substituting words with feasible alternatives, leading to easily distinguishable, implausible sentences. Consequently, there is a marked difference in similarity scores. For stable training, an upper bound $u$ on the threshold is crucial:
{
\small
\begin{equation}
    Th_k^{t}=min\left(u,\frac{1}{N}\sum_{(I,T) \in \mathcal{B}}(S^{t-1}(I,T)-S^{t-1}(I,T_k)))\right)
    \label{upperbound}
\end{equation}
}
Subsequently, incorporating the loss weight hyper-parameters $\alpha$ and $\beta$, the final loss function can be expressed as follows:
{
\small
\begin{equation}
    \mathcal{L}=\mathcal{L}_{itc(hn)}+\alpha \cdot \mathcal{L}_{imc} +\beta \cdot \mathcal{L}_{cmr}
\end{equation}
}

\section{Experiments}

We assess the performance of our method using two models. Firstly, we employ CLIP \cite{radford2021learning}, a foundational model in the vision-language domain. Additionally, we experiment with X-VLM \cite{zeng2022multigrained}, a resilient model trained on multi-grained objectives, known for its notable performance in compositional understanding \cite{bugliarello2023measuring}.

\subsection{Setup}
\paragraph{Training}
We refer to the CLIP finetuned with our proposed losses as the \textbf{C}ompositional \textbf{E}nhanced CLIP (CE-CLIP). We train in two configurations: (1) CE-CLIP, using only the COCO dataset \cite{mscoco}, for direct comparison with NegCLIP \cite{yuksekgonul2022and}, and (2) CE-CLIP+, which leverages a combined dataset of COCO, CC3M \cite{Sharma2018ConceptualCA}, and Visual Genome \cite{visualgenome} aiming for heightened performance.

We employ the \textit{CLIP-VIT/32-B} from the Open-CLIP implementation and the \textit{X-VLM-16M} from its primary code repository for evaluation purposes.\footnote{\url{https://github.com/mlfoundations/open_clip}} Both models undergo fine-tuning over 5 epochs following previous works \cite{yuksekgonul2022and, doveh2022teaching} using 2 A100 GPUs. We allocate batch sizes of 256 for CLIP and 64 for X-VLM fine-tuning. All training parameters, like learning rate, decay rate, etc., remain at default values. We conducted a hyper-parameter search for $\alpha,\beta$ with optimal values of $\alpha=0.2$ and $\beta=0.4$.
\vspace{-2mm}
\paragraph{Evaluation}
We evaluate our method on several vision-language(vl)-compositional benchmarks: ARO\cite{yuksekgonul2022and}, VL-CheckList\cite{zhao2022vlchecklist}, VALSE\cite{parcalabescu-etal-2022-valse}, and SugarCrepe\cite{hsieh2023sugarcrepe} (bias-mitigated version of CREPE\cite{ma2022crepe}). Although Winoground was designed to test compositional reasoning, \citet{diwan2022winoground} highlights other challenges posed by this dataset, like commonsense reasoning and unique image/text understanding. As these are not focus of our work, we excluded Winoground from our evaluations. We evaluate our methods in zero-shot settings. Each evaluation involves classifying positive and negative captions for a given image, with a random success probability of 50\%.

For a comprehensive evaluation, we selected robust baselines:
(1) Cutting-edge generative vision-language models such as BLIP \cite{blip}, BLIP2 \cite{blip2}, and MiniGPT-4 \cite{zhu2023minigpt4};
(2) High-performing vision-language understanding models like BEIT3 \cite{beit3}, ALBEF \cite{albef}, UNITER \cite{chen2020uniter}, CyCLIP \cite{goel2022cyclip}, and X-VLM \cite{zeng2022multigrained};
(3) Compositional improvement methods such as syn-CLIP \cite{syn} and CLIP-SGVL \cite{SGVL} (both leveraging scene graph annotations), DAC \cite{DAC} (utilizing segmentation models and LLMs), and NegCLIP \cite{yuksekgonul2022and} and CLIP-SVLC \cite{doveh2022teaching} that employ hard negative.

\subsection{Compositional reasoning enhancement}

\begin{table*}[!]
\centering 
\resizebox{0.95\textwidth}{!}{
\begin{tabular}{llcccccccccccc}
\toprule
 \multirow{3}{*}{\bf Model}& \multirow{3}{*}{\bf \#Params} &\multicolumn{2}{c}{\textbf{ ARO}} &\multicolumn{10}{c}{\bf VALSE } \\
 \cmidrule(rr){3-4} \cmidrule(rrrrrrrrrr){5-14}
&  & \multirow{2}{*}{\bf Relation} & \multirow{2}{*}{\bf Attribute} & \bf Existence &\bf Plurality   &\bf Counting &\bf Sp.rel.& \multicolumn{2}{c}{\bf Actions}&\multicolumn{2}{c}{\bf Coreference}&\multirow{2}{*}{\bf  Foil-it!} &\multirow{2}{*}{\bf  Avg.}    \\
 & & & & quantifiers&number& &relations&repl.&actant swap&standard&clean\\
\midrule
Random Chance  & \multicolumn{12}{c}{50}\\ \midrule
BLIP\cite{blip} & 583M & 59.0 & 88.0 & 86.3&73.2&68.1&71.5&77.2&61.1&53.8&48.2&93.8&70.0 \\

BEIT3\cite{beit3} & 1.9B& 60.6 & 74.6  &77.4 & 74.6& 68.8&74.0&86.7&65.2&50.0&44.2&96.0&70.4 \\
BLIP2\cite{blip2} & 3.4B & $41.2^\dag$ & $71.3^\dag$  &55.5&71.5&66.0&62.4&83.6&51.6&48.6&51.9&95.9 & 65.4 \\
%from dac
MiniGPT-4\cite{blip2} & >9B & $46.9^\dag$ & $55.7^\dag$  &65.5&72.5&67.4&68.4&83.2&58.8&52.6&51.0&95.8 & 68.4 \\
%from An Examination of the Compositionality of Large Generative Vision-Language Models

\midrule
% \multicolumn{10}{c}{\multirow{2}{*}{\textit{OpenAI's VIT-B/32 CLIP}} }\\
\multicolumn{4}{l}{\textit{Scene Graph relied method}} & &  \\
syn-CLIP\cite{syn}$\dag$ &151M& 71.4 & 66.9  & -& -&-&--&-&-&-&-&-\\

% syn-CyCLIP\cite{syn}$\dag$ &151M& 69.0& 63.7 & -& -&-&-&-&-&-&-&-&-\\ 
\midrule
\multicolumn{4}{l}{\textit{Segmentation \& LLM relied method}} & & & & &  & & &\\

DAC-LLM\cite{DAC}$\dag$&151M & 81.3& 73.9 & -& -&-&-&-&-&-&-&-&-\\
DAC-SAM\cite{DAC}$\dag$ &151M& 77.2&70.5 & -& -&-&-&-&-&-&-&-&-\\ \midrule
\multicolumn{4}{l}{\textit{Hard Negative based method}} & & & & &  & & &\\
XVLM-coco\cite{zeng2022multigrained}&216M& 73.4 & 86.8 & 83.0&75.6&67.5&70.2&73.8&68.6&46.4&49.6&94.8&69.5 \\
\rowcolor{lightblue}
 CE-XVLM &216M& $\bf 73.9_{\textcolor{mygreen}{+0.5}}$  & $\bf 89.3 _{\textcolor{mygreen}{+2.5}}$ & 83.5&72.8&72.1&68.7&71.8&69.1&51.0&46.8&93.8 &$\bf 70.8_{\textcolor{mygreen}{+1.3}}$\\
\hdashline
CLIP\cite{radford2021learning}& 151M & 59.3 & 62.9 & 68.7&57.1&61.0&65.4&77.8&71.8&54.1&51.0&89.8&65.3\\
CyCLIP\cite{goel2022cyclip}$\dag$ &151M& 59.1 & 65.4 & 69.3 & 58.3&61.0&66.4&78.1&72.0&53.2&51.6&88.8&65.5 \\
SDS-CLIP\cite{sdsclip} $\dag$ &151M& 53.0 & 62.0  & -& -&-&-&-&-&-&-&-&- \\

NegCLIP\cite{yuksekgonul2022and}&151M & 80.2 & 70.5 & 76.8&71.7&65.0&72.9&81.6&84.7& 58.6&53.8&91.9&71.6\\
CLIP-SVLC\cite{doveh2022teaching} $\dag$&151M & 80.61 & 73.03 & -&-&-&-&-&-&-&-&-&-\\
\rowcolor{lightblue}
CE-CLIP&151M & $ 83.0_{\textcolor{mygreen}{+23.7}}$ & $ 76.4_{\textcolor{mygreen}{+13.5}}$ &  78.6& 77.7&64.4& 74.4&81.2& 88.6&54.7& 54.8& 93.7& $ 72.5_{\textcolor{mygreen}{+7.2}}$\\
\rowcolor{lightblue}
CE-CLIP+&151M & $\bf 83.6_{\textcolor{mygreen}{+24.3}}$ & $\bf 77.1_{\textcolor{mygreen}{+14.2}}$ &  84.5& 79.2&67.8& 76.4&83.4& 89.4&56.7& 57.8& 94.7& $\bf 76.7_{\textcolor{mygreen}{+11.4}}$\\

\bottomrule

 \end{tabular}}

\captionsetup{font=small}
\caption{\textbf{Results (\%) on ARO and VALSE}. The best scores for each section are highlighted in bold. $\dag$ represents scores are extracted from papers. Empty scores suggest that the model's codebase has not been released.}
\label{tab:mainresult}
\end{table*}

\begin{figure*}[h]
\centering

% First table
\begin{minipage}[c]{0.65\textwidth}
    \centering
    \resizebox{\linewidth}{!}{
    \begin{tabular}{lcccccccccc}
    \toprule
     \multirow{2}{*}{\bf Model} &\multicolumn{4}{c}{\bf REPLACE} &\multicolumn{3}{c}{\bf SWAP }&\multicolumn{3}{c}{\bf ADD } \\
     \cmidrule(rrrr){2-5}  \cmidrule(rrr){6-8}\cmidrule(rrr){9-11}
     & Object&Attribute&Relation& Avg & Object&Attribute& Avg& Object&Attribute& Avg
        \\
    \midrule
    Human & 100& 99& 97& 98.7 & 99& 100&99.5& 99& 99&99\\ \midrule
    Vera\cite{liu2023vera} & 49.4 & 49.6 & 49.1 &49.4  &49.4 & 49.2& 49.3 &49.4 & 49.6& 49.5\\ 
    Grammar\cite{grammar} & 50&50&50& 50& 50& 50& 50& 50& 50& 50\\ 
    BLIP2\cite{blip2} $\dag$ & - & - & - & 86.7 & - & - & 69.8 &-&-&86.5 \\\midrule
    CLIP & 90.9& 80& 69.2& 80.2& 61.4& 64&62.7& 77.2& 68.2&72.7\\
    NegCLIP & 92.7& 85.9& 76.5&85.0&  75.2& 75.4&75.3& 88.8& 82.8&85.8\\
    \rowcolor{lightblue}
    CE-CLIP &  93.1&88.8& 79&$\bf 87.0_{\textcolor{mygreen}{+6.8}}$& 72.8& 77&$74.9_{\textcolor{mygreen}{+12.2}}$& 92.4& 93.4&$\bf 92.9_{\textcolor{mygreen}{+20.2}}$\\
    \rowcolor{lightblue}
    CE-CLIP+ &  93.8&90.8& 83.2&$\bf 89.3_{\textcolor{mygreen}{+9.1}}$& 76.8& 79.3&$\bf78.0_{\textcolor{mygreen}{+15.3}}$& 93.8& 94.9&$\bf 94.4_{\textcolor{mygreen}{+21.7}}$\\
    \bottomrule
     \end{tabular}}
    \captionsetup{font=small}
    \captionof{table}{\textbf{Results(\%) on SugarCrepe}. Vera and Grammar are text-only models.}
    \label{tab:sugarcrepe}
\end{minipage}
\vspace{-2mm}
\hfill
% Second table
\begin{minipage}[c]{0.33\textwidth}
    \centering
     \includegraphics[width=\linewidth]{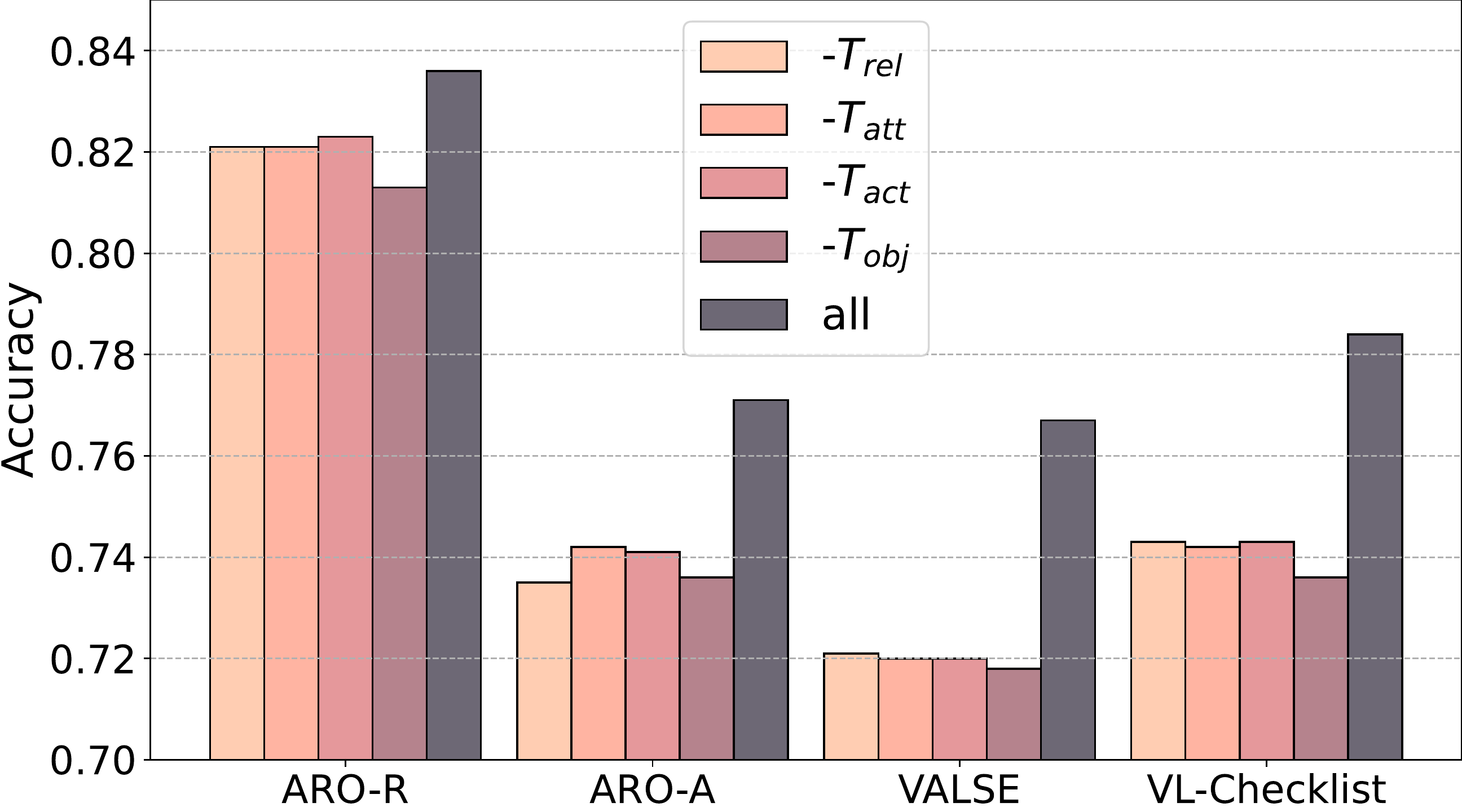}
     \captionsetup{font=small}
    \caption{Ablations on hard-negative types}
    \label{tab:hardnegative types}
\end{minipage}

\end{figure*}

\begin{table*}[!tb]
\centering 
\resizebox{0.95\textwidth}{!}{
\begin{tabular}{llcccccccccccccc}
\toprule
 \multirow{2}{*}{\bf Model}& \multirow{2}{*}{\bf \#Params} & \multicolumn{6}{c}{\bf Attribute} & \multicolumn{3}{c}{\bf Object}  & \multicolumn{3}{c}{\bf Relation} & \multirow{2}{*}{\textbf{Avg}}\\ \cmidrule(rrrrr){3-8} \cmidrule(rrr){9-11}\cmidrule(rrr){12-14} &
    & Action & Color & Material & Size & State & Avg & Location & Size & Avg & Action & Spatial&  Avg     \\
    \midrule
    Random Chance &  \multicolumn{13}{c}{50}\\ \midrule
        ALBEF\cite{albef} $\dag$ & 210M&  81.7 & 84.2 & 87.3&69.5&72.08& 79.3 & 81.7 & 80.5 & 81.1 & 70.5 & 64.6 & 66.5 &  75.6\\
        UNITER\cite{chen2020uniter}$\dag$& 300M & 72.6 & 76.2 & 75.8 & 63.5 &68.1 & 71.3 & 82.4 & 81.5 & 81.9 & 69.2& 61.5&64.7&72.6\\
        BLIP\cite{blip}$\dag$ &583M &79.5&83.2&84.7&59.8&68.8&75.2&83.0&81.3&82.2&59.5&75.7&70.5&75.7\\
        BEIT3\cite{beit3}& 1.9B& 79.6 & 78.5 &80.1 &63 &68.4  &73.9 &85.2 &83.8 &84.5 &76.6 &62.3 & 69.4 & 75.3\\
        BLIP2\cite{blip2}$\dag$ &3.4B &81.0& 86.2& 90.3& 61.7& 70.1&77.8&85.4&84.3&84.9&84.9&56.2&70.6& 77.8\\
        MiniGPT-4\cite{zhu2023minigpt} $\dag$ & >9B& - & - & - & - & - & 71.3 & -  & - & 84.2 & 84.1 & - & -&- \\
   \midrule
   \multicolumn{4}{l}{\textit{Scene Graph relied method}} & &  \\
    CLIP-SGVL\cite{SGVL}$\dag$ &>151M & 76.6& 78.0 & 80.6 & 59.7 & 61.2 & 71.2 & 83.0 & 81.3 & 82.6 & 79.0 & - & -&-\\
    syn-CLIP\cite{syn} $\dag$ &151M & - &- &- &- &- & 70.4 & - & - & -& -&-& 69.4&- \\
    \midrule
    \multicolumn{4}{l}{\textit{Segmentation \& LLM relied method}} & & &   \\
DAC-LLM\cite{DAC}$\dag$&151M & - & -& -&-&- & 77.3 & - & - & 87.3 & 86.4 &- & -& -\\
DAC-SAM\cite{DAC}$\dag$ &151M& - & -& -&-&- & 75.8 & - & - & 88.5 & 89.8 &- & -& - \\ \midrule
   \multicolumn{4}{l}{\textit{Hard Negative based method}} & & & & &  & & &\\
    XVLM-coco\cite{zeng2022multigrained} &216M &80.4&81.1&83.1&60.3&70.8&\textbf{75.1}&86.3&85.3&85.8&79.0&61.8&70.4&76.5\\
    \rowcolor{lightblue}
      CE-XVLM&216M & 80.5&76.0&80.6&67.2&69.8&$74.8_{\textcolor{red}{-0.3}}$&87.3&86.6&$ \bf 86.9 _{\textcolor{mygreen}{+1.1}}$&80.8&78.6&$\bf 79.7_{\textcolor{mygreen}{+9.3}}$&$\bf 78.6_{\textcolor{mygreen}{+2.1}}$\\ 
   \hdashline
    CLIP\cite{radford2021learning}&151M  & 70.5& 69.4& 69.5 & 60.7 & 67 & 67.4& 80.2 & 79.7& 80.0 & 72.2 & 53.8 & 63.0 &  69.2 \\ 
     CLIP-SVLC\cite{doveh2022teaching}$\dag$ &151M & 69.4&77.5& 77.4&73.4&62.3& 72.0 & - & - & 85.0 & 74.7&63.2 &68.95& 74.2\\

     NegCLIP\cite{yuksekgonul2022and} &151M  &  72.1&75.7&78.1&61.3&67.3&70.9&84.4&83.8&84.1&80.7&57.1&68.9&73.4\\
     \rowcolor{lightblue}
     CE-CLIP&151M & 75.6&72.7&79.7&65.3&69.8&$72.6_{\textcolor{mygreen}{+5.2}}$&84.8&84.5&$ 84.6_{\textcolor{mygreen}{+4.6}}$&78.5&65.0&$ 71.8_{\textcolor{mygreen}{+8.8}}$&$ 75.1_{\textcolor{mygreen}{+5.9}}$ \\ 
    \rowcolor{lightblue}
    CE-CLIP+&151M & 78.5&83.5&85.2&65.8&70.8&$\bf 76.7_{\textcolor{mygreen}{+9.3}}$&86.7&85.9&$\bf 86.3_{\textcolor{mygreen}{+6.3}}$&81.0&68.4&$\bf 74.7_{\textcolor{mygreen}{+11.7}}$&$\bf 78.4_{\textcolor{mygreen}{+9.2}}$ \\

\bottomrule
\end{tabular}}
\captionsetup{font=small}
\caption{\textbf{Results (\%) on VL-CheckList.}  The best scores for each section are highlighted in bold. $\dag$ represents scores are extracted from papers. Empty scores suggest that the model's codebase has not been released.}
\label{tab:mainresult_vlchecklist}
\end{table*}
We present results for ARO and VALSE in \cref{tab:mainresult}, VL-CheckList in \cref{tab:mainresult_vlchecklist}, and SugarCrepe in \cref{tab:sugarcrepe}. Our CE-CLIP model, which is trained on the same dataset as NegCLIP, surpasses all methods utilizing hard negatives across all benchmarks. It demonstrates significant improvements over the baseline CLIP model: 23.7\% on ARO-Relation, 13.5\% on ARO-Attribute, 7.2\% on VALSE, 5.2\% on VL-CheckList, and 12.1\% on SugarCrepe. This indicates that our approach more effectively utilizes hard negatives through intra-modal contrasting and cross-modal ranking. Notably, the smallest absolute improvement was observed on VL-CheckList, likely because %it is the largest benchmark, encompassing a comprehensive evaluation of various aspects of compositional understanding and integrating diverse datasets. 
this benchmark presents an out-of-distribution challenge for our CE-CLIP, given that it is only fine-tuned on COCO, while VL-CheckList integrates several diverse datasets. Conversely, we note a substantial improvement on the ARO benchmark, which could be attributed to the hard negative types in our model that are specifically tailored to enhance the understanding of objects and attributes. Additionally, the significant gains observed on SugarCrepe, a benchmark designed to mitigate language bias in other benchmarks and provide a more accurate reflection of a model's compositional understanding, are particularly noteworthy.

The CE-CLIP+, trained on a more comprehensive dataset, achieves superior performance, with an average improvement of 24.3\% and 14.2\% on ARO Relation and Attribution splits, 11.4\% on VALSE, 9.2\% on VL-CheckList, and 14.4\% on SugarCrepe which translates to an impressive average accuracy of 87.5\%. Similar to CE-CLIP, the greatest and smallest improvements were observed in ARO and VL-CheckList, respectively, reinforcing our initial hypothesis. The out-of-distribution challenge observed in CE-CLIP has been substantially mitigated in CE-CLIP+ through training on a varied range of data distributions. For example, in the Attribute evaluation split, CE-CLIP showed a modest 5.2\% improvement on VL-CheckList and a significant 13.5\% increase on ARO. Impressively, CE-CLIP+ outperforms CE-CLIP by 0.7\% (76.4$\rightarrow$77.1) on ARO-Attribute and an exceptional 4.1\%  (72.6$\rightarrow$76.7) on the VL-CheckList Attribute split. This underscores the challenges of out-of-distribution evaluation encountered by CE-CLIP and illustrates the effectiveness of augmenting dataset size as a remedy. Overall, CE-CLIP+ demonstrates robust performance, surpassing models with significantly larger parameters or those trained with extra resources and annotations across the majority of benchmarks. This strengthens the potential scalability of our method within extensive pre-training frameworks, although we acknowledge the necessity for further investigation.

X-VLM shows a modest improvement compared with large improvement gained on CE-CLIP, primarily due to differences in pretraining approaches. X-VLM is pretrained on multiple fine-grained tasks that necessitate specific object bounding box annotations, whereas CLIP is trained directly on automatically crawled, noisy image-text pairs. Our simple annotation-free method can bolster the already strong X-VLM, further emphasizing its distinctive characteristics in learning compositionality. However, our method is most beneficial for CLIP like models that do not already benefit from object annotations during pre-training.

% \subsection{Adaptive threshold imply Curriculum}
\subsection{Emergence of curriculum learning}
% The adaptive threshold value curve (\cref{fig:threhsold_loss} d) shows a sharp increase for \textit{Threshold Relation} value, primarily because the relation-swap hard negatives can generate sentences with semantic and grammatical errors (e.g., sentences in \cref{fig:hardnegative}). This enables the model to easily distinguish between authentic and hard negative captions \cite{hsieh2023sugarcrepe}. Accordingly, our approach increases the task's complexity by elevating the threshold, thereby providing a stronger supervisory signal to the model, forcing it to identify a larger difference between these captions.

% The threshold value is determined by the average disparity between true and hard negative similarity scores, representing the task's complexity and the model's capacity to differentiate various hard negatives. As depicted in the loss curve (\cref{fig:threhsold_loss} b), the CMR loss reaches stability after early fluctuations, signifying a balance between the escalating complexity of the task and the model's evolving adaptability. This equilibrium highlights the curriculum learning feature integral to our proposed adaptive threshold approach.

In this section, we illustrate how the adaptive threshold in the cross-modal loss facilitates curriculum learning during fine-tuning. We analyze the evolution of the threshold values and losses over time, with the curve in \cref{fig:threhsold_loss}d showing a sharp increase in the \textit{Threshold Relation} value. This rise is mainly due to the semantic and grammatical errors in relation-swap hard negatives (e.g., sentences in \cref{fig:hardnegative}), simplifying the model's task of differentiating authentic captions from hard negatives. Consequently, the elevated threshold counters this by increasing the task difficulty, providing a stronger supervisory signal and compelling the model to discern greater differences between these captions.

The threshold, calculated as the average gap between true and hard negative similarity scores, mirrors the task’s complexity and the model’s discernment capability. CE-CLIP+'s training loss curve (\cref{fig:threhsold_loss}b) indicates that CMR loss stabilizes after initial fluctuations, striking a balance between escalating task difficulty and the model’s adaptive capacity, thereby highlighting the inherent curriculum learning.

% The emergence of curriculum learning, yields satisfactory outcomes without the need for extensive hyper-parameter tuning. In contrast, using adaptive thresholds and exploring \( n \) different values for each of the four thresholds would require an impractical \( n^4 \) trials. Consequently, we utilized a single value for the hard negatives threshold in all cases and compared this fixed threshold strategy. \cref{fig:threhsold_loss} a shows the average results of CE-CLIP+ over 5 benchmarks using various threshold strategies. The adaptive threshold strategy outperforms the fixed strategy and converges faster after the initial epoch. We hypothesize that initially, when the threshold is low, the model receives a smaller supervision signal compared with fixed threshold strategy. As training progresses, the adaptive strategy adjusts the threshold in response to the task’s complexity and the model’s capacity. This enhances learning efficiency and generalization, allowing the model to optimize more smoothly and avoid local minima.

The emergence of curriculum learning achieves satisfactory outcomes without needing extensive hyper-parameter tuning. In contrast, a fixed threshold strategy would require impractical \( n^4 \) trials for exploring \( n \) different values across four thresholds. \cref{fig:threhsold_loss}a compares CE-CLIP+ results across 5 benchmarks using various thresholds, showing adaptive approach outperforms the fixed ones and converges faster. Initially, the adaptive strategy provides a smaller supervision signal compared to the fixed approach but as the training progresses, it adjusts the threshold according to the task complexity and model capacity. This adjustment enhances learning efficiency and generalization.
% allowing smoother optimization and minimizing the risk of local minima.

\subsection{Ablation studies}

We present ablation studies to understand the effectiveness of different components of our method. We conduct these ablations using our best model CE-CLIP+.
\vspace{-4mm}

\begin{table}
    \centering
    \resizebox{\linewidth}{!}{
        \begin{tabular}{lcccccccc} \toprule
            \textbf{Model}  & \textit{itc(hn)}  & \textit{IMC}&\textit{CMR}& \textbf{ARO-R}& \textbf{ARO-A} & \textbf{VALSE} &  \textbf{VLCheckList}  & \textbf{Avg} \\ \midrule
            CLIP  & & && 59.3 & 62.9 & 67.0 & 69.2  & 64.6\\
            & \checkmark  & && 81.6 & 72.0 & 74.2&73.6  & 75.4\\
            &  \checkmark& \checkmark& & 82.6 & 75.8 & 75.9 & 76.6  &77.7\\
            &  \checkmark & &\checkmark& 82.3 & 72.6 & 75.5 & 77.8  &77.1\\
             \rowcolor{lightblue}
           CE-CLIP+ & \checkmark &\checkmark &\checkmark & \textbf{83.6} & \textbf{77.1} & \textbf{76.7} & \textbf{78.4}  &\textbf{79.0}\\
            \bottomrule
        \end{tabular}}
        \captionsetup{font=small}
        \caption{\textbf{Ablation of losses.} \textit{itc(hn)} represents image-text contrastive with additional hard negatives.}
        \label{tab:ablation}
 
\end{table}
\begin{figure}[]
    \centering
     
    \label{fig:derivation}
    \includegraphics[width=\linewidth]{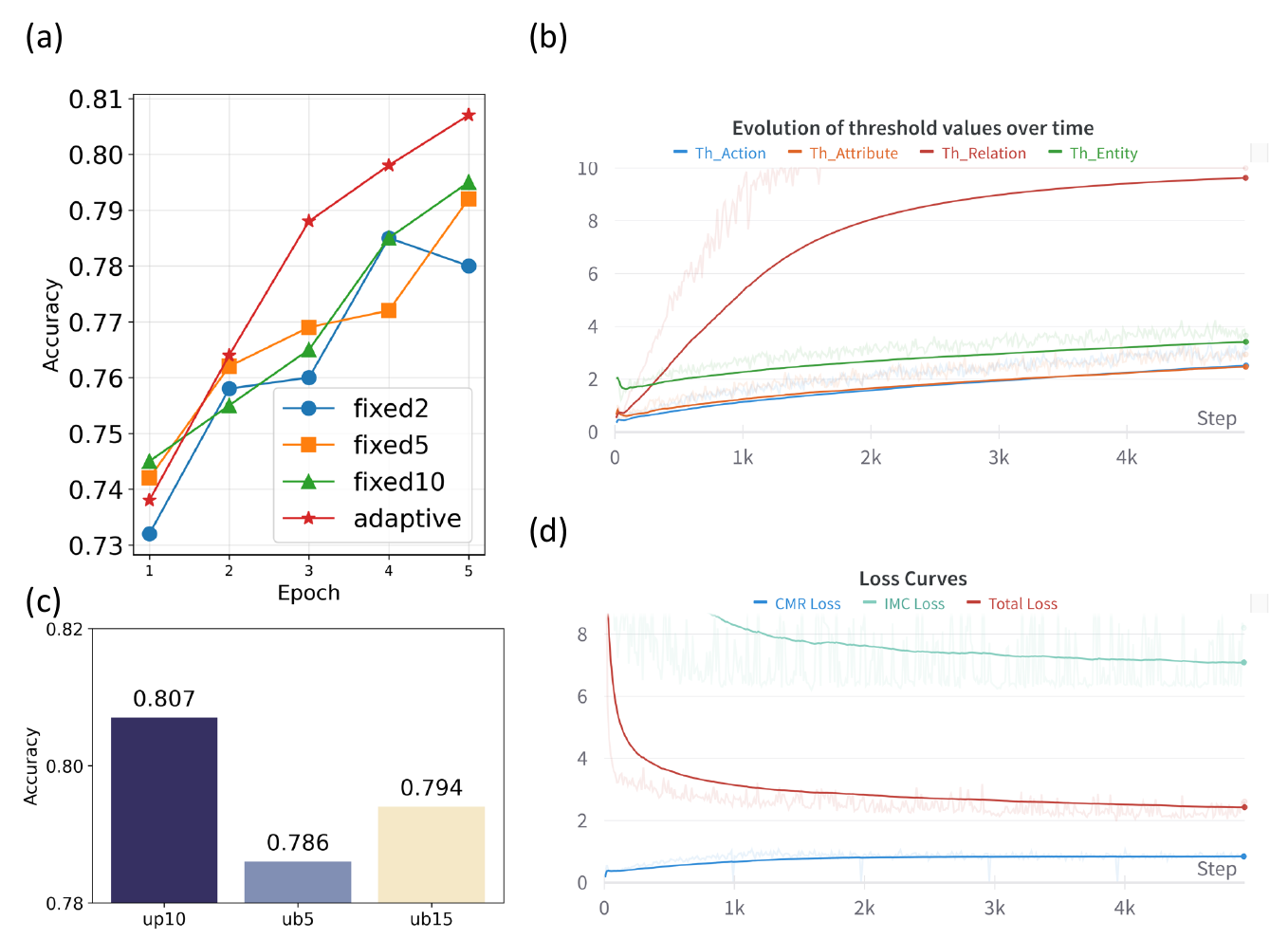}
      \captionsetup{font=small}
     \caption{\textbf{Ablation studies.} (a) Adaptive vs Fixed threshold with values 2, 5, 10; (b) Evolution of threshold over time ; (c) Performance with different upper bounds on threshold. (d) Loss curves showing stabilization of the CMR loss after initial training steps.}
     \label{fig:threhsold_loss}
  
\end{figure}

\begin{figure*}[h]
    \begin{minipage}[]{0.25\linewidth}
        \centering
        \includegraphics[width=\linewidth]{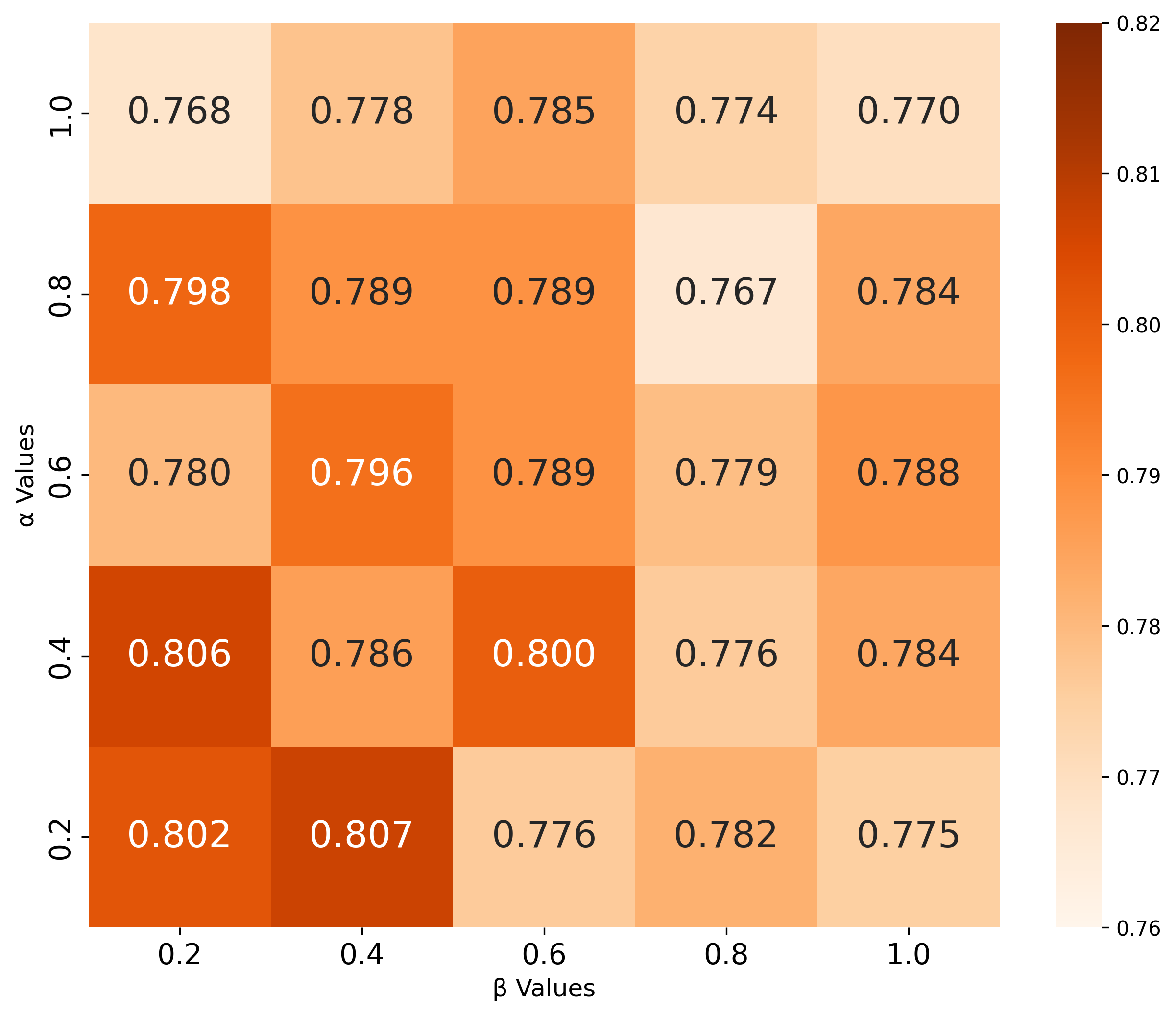}
        \captionsetup{font=small}
        \caption{\textbf{Ablations of $\alpha,\beta$.} }
        \label{fig:alphabeta}
            
    \end{minipage}%
    \hfill
    \begin{minipage}[]{0.32\linewidth}
        \centering
        \resizebox{\linewidth}{!}{
        \begin{tabular}{lcccc}
        \toprule
        Task & CLIP& CLIP-FT & CE-CLIP & CE-CLIP+ \\
        \midrule
        \multicolumn5{c}{\textit{compositional tasks}}\\\midrule
        \textbf{ARO-R\cite{yuksekgonul2022and}} & 59.3 & 61.7 & 83& 83.6(\textcolor{mygreen}{+24.3})  \\
        \textbf{ARO-A\cite{yuksekgonul2022and}} & 62.9 & 66.1 & 76.4& 77.1(\textcolor{mygreen}{+14.2})\\
        \textbf{VALSE\cite{parcalabescu-etal-2022-valse}} & 65.3 &71.8& 72.5&76.7(\textcolor{mygreen}{+11.4}) \\
        \textbf{VLChecklist\cite{zhao2022vlchecklist}} & 69.2 & 68.6 &75.1 & 78.4(\textcolor{mygreen}{+9.2}) \\
        \textbf{SugarCrepe\cite{hsieh2023sugarcrepe}} & 73.1& 77.2 &85.2&87.5(\textcolor{mygreen}{+14.4})\\
        \midrule
        \multicolumn{5}{c}{\textit{standard tasks}}\\\midrule
        \textbf{T2I R@5\cite{coco2015}} & 56.0 &66.2 & 69.4& 72.3(\textcolor{mygreen}{+13.4})\\
        \textbf{I2T R@5\cite{coco2015}} &  75.0 &78.3 &74.3 & 76.1(\textcolor{mygreen}{+1.1}) \\
        % resutls from https://github.com/LAION-AI/CLIP_benchmark/blob/main/benchmark/results.ipynb
        \textbf{ImageNet1K} &  93.2& 92.8 &92.6 & 92.7(\textcolor{red}{-0.5}) \\
        \textbf{CIFAR10} &  94.2& 94.2 &93.8 & 93.8(\textcolor{red}{-0.4}) \\
        \textbf{CIFAR100} &  79.0& 79.1 &78.0 & 78.1(\textcolor{red}{-0.9}) \\
        
        \bottomrule
        \end{tabular}}
        \captionsetup{font=small}
        \captionof{table}{Performance on standard image-text retrieval and image classification. Improvements in green are calculated w.r.t CLIP.}
        \label{tab:cocoir}
    
    \end{minipage}
     \hfill
      \begin{minipage}[]{0.4\linewidth}
        \centering
        \includegraphics[width=\linewidth]{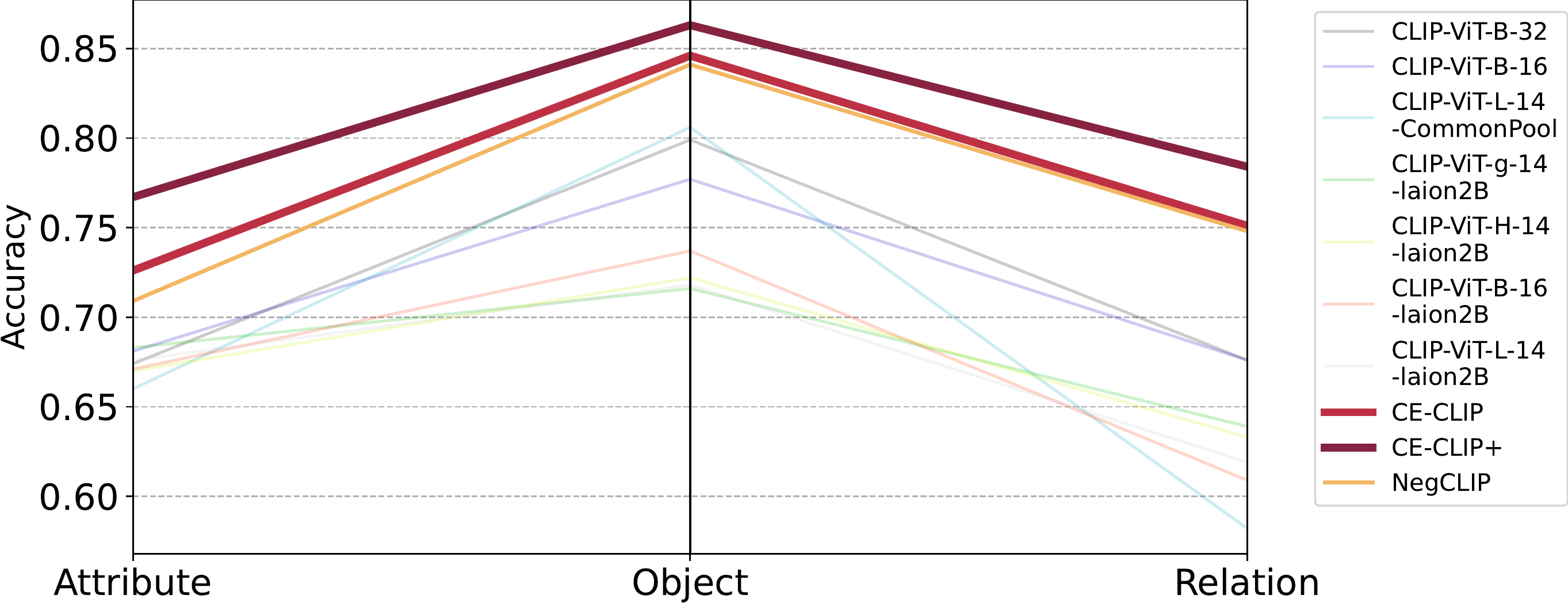}
        \captionsetup{font=small}
        % \caption{Performance of different sized CLIP on VL-CheckList}
        \caption{Impact of scaling-up the model on VL-CheckList performance.}
        \label{fig:vl_type_ab}
    \end{minipage}%

\end{figure*}

\paragraph{Losses.}
The impact of each proposed loss is detailed in Tab \ref{tab:ablation}. Notably, the introduction of hard negatives led to tremendous performance gains, highlighting their pivotal role in contrastive learning. Each individual loss we introduced showed significant improvements as well across all benchmarks. The best performance is achieved when all losses are combined, thus demonstrating the effectiveness of our approach. 
% We achieved peak performance across all benchmarks with combined losses.

\vspace{-3mm}

\paragraph{Hard Negative Types} As shown in \cref{tab:hardnegative types}, 
% hard-negative types are crafted to focus on different aspects of compositionality.
each type of hard-negative uniquely benefits the model, the object hard negatives benefitting the most. Combining all types yields the best results. The success of our flexible approach indicates that incorporating additional types, such as numerical negatives \citep{paiss2023teaching}, may further boost performance.

\vspace{-2mm}
\paragraph{Upper Bound on Threshold.} 

Setting a threshold upper bound prevents training collapse. Our ablation study, as detailed in Fig.~\ref{fig:threhsold_loss}c, demonstrates that an upper bound of 10 yields optimal performance by effectively constraining the maximum value of the \textit{Threshold Relation} (Fig.~\ref{fig:threhsold_loss}b), thereby ensuring stability during the training process.

\vspace{-3mm}

\paragraph{Loss Weight.} \cref{fig:threhsold_loss}d shows the divergence of CMR loss scale from IMC loss, highlighting the importance of proper loss weight selection for training. \cref{fig:alphabeta} reveals that our method is robust across 5 benchmarks with varying $\alpha$ and $\beta$ values, though larger $\alpha$ and $\beta$ decrease performance. Optimal outcomes occur at $\alpha=0.2$ and $\beta=0.4$.

\subsection{Performance on standard benchmarks}

% To analyze how method affect standard vision-language performance we conduct experiments on COCO retrieval and ELEVATER image classification \cite{li2022elevater}. Prior studies indicate that progress in compositional understanding can negatively impact the performance on standard image-text retrieval and image classification \cite{doveh2022teaching, yuksekgonul2022and}. Results in \cref{tab:cocoir} show both improve text-to-image retrieval with minimal impact on classification. Emphasizing compositional understanding, both enhance performance across all benchmarks. To demonstrate that the results on standard image-text retrieval are not inflated due to fine-tuning on COCO, we also report results from a variant of CLIP that was fine-tuned on COCO, which we refer to as "CLIP-FT". Our findings indicate that CE-CLIP and CE-CLIP+ outperform CLIP-FT in text-to-image retrieval tasks, while showing a slight underperformance in image-to-text retrieval. We hypothesize that this may be attributed to our method's exclusive use of textual hard negatives.

Previous studies \cite{doveh2022teaching, yuksekgonul2022and}, suggest that advancements in compositional understanding might negatively affect performance on standard image-text retrieval and image classification tasks. To investigate this, we evaluate our method on zero-shot image-text retrieval on COCO and linear probing on ImageNet-1k \cite{imagenet} and CIFAR \cite{cifar}.
As shown in \cref{tab:cocoir}, our results demonstrate improvements in text-to-image retrieval with minimal impact on image classification accuracy. By prioritizing compositional understanding, our CE-CLIP and CE-CLIP+ enhance performance across all evaluated benchmarks. Furthermore, to demonstrate that our enhancements in COCO image-text retrieval are not merely a result of fine-tuning on COCO, we include comparative results from CLIP-FT, a COCO fine-tuned variant of CLIP. Our findings indicate that both CE-CLIP and CE-CLIP+ outperform CLIP-FT in text-to-image retrieval, albeit with a slight underperformance in image-to-text retrieval. We hypothesize this could be due to our method's exclusive reliance on textual hard negatives.

% We also investigate whether CE-CLIP retains the original CLIP's visual strengths when applied to the Visual Question Answering task. This is done by integrating CE-CLIP into MAPL\cite{mañas2023mapl} and training it on the COCO dataset, followed by conducting a zero-shot evaluation on VQAv2 \cite{vqav2}. The results show that CE-CLIP, with an accuracy of 39.82, performs comparably to the original CLIP, which has an accuracy of 39.78.  This performance suggests that CE-CLIP retains the visual capacity of the original model.

We investigated the impact of integrating CE-CLIP into MAPL \cite{mañas2023mapl} on the Visual Question Answering (VQA) task by training on the COCO dataset and conducting a zero-shot evaluation on VQAv2 \cite{vqav2}. The findings indicate that CE-CLIP, achieving an accuracy of 39.82, closely matches the original CLIP's performance of 39.78. This demonstrates that CE-CLIP preserves the visual strengths of CLIP.

\subsection{Can scaling-up alone solve compositionality}

To substantiate our assertion in \cref{fig:motivation} that standard contrastive learning as implemented in CLIP fails to grasp compositionality, we tested several scaled-up versions of CLIP models including the largest ViT-G/14 trained on LAION-2B from Open-CLIP, on the VLChecklist benchmark. As \cref{fig:vl_type_ab} shows, none of these scaled-up models surpass our base-sized CE-CLIP model. This shows that scaling-up the model alone is not enough for comprehending compositionality, underscoring the significance of our work and the need for more research in this field.

\vspace{-2mm}
\section{Conclusion}
Our study addresses the challenge of compositional understanding in VLMs  , we expand image-text contrastive loss and introduce two losses that infuse compositional supervision into pretrained VLMs using a featured hard negative generation strategy. Our intra-modal contrastive loss mitigates high intra-modal similarity while our cross-modal rank loss ensures a minimum semantic distance between true and hard negative image-text pairs, with the adaptive threshold functioning as curriculum learning to enhance performance. Empirically, our method achieves superior performance in 5 compositional benchmarks, surpassing previous methods without requiring additional annotations or resources. Scaling the dataset size further boosts performance, highlighting our method's potential for VLMs and its promise for broader applications and capabilities.

% Our study addresses the challenge of compositional understanding in VLMs by augmenting image-text contrastive loss with two novel losses, leveraging a hard negative generation strategy. The intra-modal contrastive loss reduces intra-modal similarity, and the cross-modal rank loss maintains semantic distance between true and hard negative pairs, using adaptive thresholding for improved performance. This approach notably outperforms others in 5 compositional benchmarks, without extra annotations or resources. Scaling the dataset size further boosts performance, highlighting our method's pretraining potential for VLMs and its promise for broader applications and capabilities.

% While each of our losses exhibits improvement, their combination essentially constitutes a multitask problem. Consequently, the issue of merging these two losses to attain stable cumulative performance is a challenge we reserve for future investigation.

{
    \small
    \bibliographystyle{ieeenat_fullname}
    \bibliography{main}

\begin{thebibliography}{80}
\providecommand{\natexlab}[1]{#1}
\providecommand{\url}[1]{\texttt{#1}}
\expandafter\ifx\csname urlstyle\endcsname\relax
  \providecommand{\doi}[1]{doi: #1}\else
  \providecommand{\doi}{doi: \begingroup \urlstyle{rm}\Url}\fi

\bibitem[Alayrac et~al.(2022)Alayrac, Donahue, Luc, Miech, Barr, Hasson, Lenc, Mensch, Millican, Reynolds, Ring, Rutherford, Cabi, Han, Gong, Samangooei, Monteiro, Menick, Borgeaud, Brock, Nematzadeh, Sharifzadeh, Binkowski, Barreira, Vinyals, Zisserman, and Simonyan]{alayrac2022flamingo}
Jean-Baptiste Alayrac, Jeff Donahue, Pauline Luc, Antoine Miech, Iain Barr, Yana Hasson, Karel Lenc, Arthur Mensch, Katie Millican, Malcolm Reynolds, Roman Ring, Eliza Rutherford, Serkan Cabi, Tengda Han, Zhitao Gong, Sina Samangooei, Marianne Monteiro, Jacob Menick, Sebastian Borgeaud, Andrew Brock, Aida Nematzadeh, Sahand Sharifzadeh, Mikolaj Binkowski, Ricardo Barreira, Oriol Vinyals, Andrew Zisserman, and Karen Simonyan.
\newblock Flamingo: a visual language model for few-shot learning, 2022.

\bibitem[Awadalla et~al.(2023)Awadalla, Gao, Gardner, Hessel, Hanafy, Zhu, Marathe, Bitton, Gadre, Sagawa, Jitsev, Kornblith, Koh, Ilharco, Wortsman, and Schmidt]{awadalla2023openflamingo}
Anas Awadalla, Irena Gao, Josh Gardner, Jack Hessel, Yusuf Hanafy, Wanrong Zhu, Kalyani Marathe, Yonatan Bitton, Samir Gadre, Shiori Sagawa, Jenia Jitsev, Simon Kornblith, Pang~Wei Koh, Gabriel Ilharco, Mitchell Wortsman, and Ludwig Schmidt.
\newblock Openflamingo: An open-source framework for training large autoregressive vision-language models, 2023.

\bibitem[Basu et~al.(2023)Basu, Sanjabi, Massiceti, Hu, and Feizi]{sdsclip}
Samyadeep Basu, Maziar Sanjabi, Daniela Massiceti, Shell~Xu Hu, and Soheil Feizi.
\newblock Augmenting clip with improved visio-linguistic reasoning.
\newblock \emph{ArXiv}, abs/2307.09233, 2023.

\bibitem[Bugliarello et~al.(2023)Bugliarello, Sartran, Agrawal, Hendricks, and Nematzadeh]{bugliarello2023measuring}
Emanuele Bugliarello, Laurent Sartran, Aishwarya Agrawal, Lisa~Anne Hendricks, and Aida Nematzadeh.
\newblock Measuring progress in fine-grained vision-and-language understanding, 2023.

\bibitem[Cascante-Bonilla et~al.(2023)Cascante-Bonilla, Shehada, Smith, Doveh, Kim, Panda, Varol, Oliva, Ordonez, Feris, and Karlinsky]{syn}
Paola Cascante-Bonilla, Khaled Shehada, James Smith, Sivan Doveh, Donghyun Kim, Rameswar Panda, G{\"u}l Varol, Aude Oliva, Vicente Ordonez, Rog{\'e}rio~Schmidt Feris, and Leonid Karlinsky.
\newblock Going beyond nouns with vision \& language models using synthetic data.
\newblock \emph{ArXiv}, abs/2303.17590, 2023.

\bibitem[Chen et~al.(2017)Chen, Chen, Zhang, and Huang]{chen2017triplet}
Weihua Chen, Xiaotang Chen, Jianguo Zhang, and Kaiqi Huang.
\newblock Beyond triplet loss: a deep quadruplet network for person re-identification, 2017.

\bibitem[Chen et~al.(2020)Chen, Li, Yu, Kholy, Ahmed, Gan, Cheng, and Liu]{chen2020uniter}
Yen-Chun Chen, Linjie Li, Licheng Yu, Ahmed~El Kholy, Faisal Ahmed, Zhe Gan, Yu Cheng, and Jingjing Liu.
\newblock Uniter: Universal image-text representation learning, 2020.

\bibitem[Cho et~al.(2022)Cho, Yoon, Kale, Dernoncourt, Bui, and Bansal]{Cho2022FinegrainedIC}
Jaemin Cho, Seunghyun Yoon, Ajinkya Kale, Franck Dernoncourt, Trung Bui, and Mohit Bansal.
\newblock Fine-grained image captioning with clip reward.
\newblock \emph{ArXiv}, abs/2205.13115, 2022.

\bibitem[Cortes and Vapnik(1995)]{cortes1995support}
Corinna Cortes and Vladimir Vapnik.
\newblock Support-vector networks.
\newblock \emph{Machine learning}, 20:\penalty0 273--297, 1995.

\bibitem[Deng et~al.(2009)Deng, Dong, Socher, Li, Li, and Fei-Fei]{imagenet}
Jia Deng, Wei Dong, Richard Socher, Li-Jia Li, Kai Li, and Li Fei-Fei.
\newblock Imagenet: A large-scale hierarchical image database.
\newblock In \emph{2009 IEEE Conference on Computer Vision and Pattern Recognition}, pages 248--255, 2009.

\bibitem[Diwan et~al.(2022)Diwan, Berry, Choi, Harwath, and Mahowald]{diwan2022winoground}
Anuj Diwan, Layne Berry, Eunsol Choi, David Harwath, and Kyle Mahowald.
\newblock Why is winoground hard? investigating failures in visuolinguistic compositionality, 2022.

\bibitem[Doveh et~al.(2022)Doveh, Arbelle, Harary, Panda, Herzig, Schwartz, Kim, Giryes, Feris, Ullman, and Karlinsky]{doveh2022teaching}
Sivan Doveh, Assaf Arbelle, Sivan Harary, Rameswar Panda, Roei Herzig, Eli Schwartz, Donghyun Kim, Raja Giryes, Rogerio Feris, Shimon Ullman, and Leonid Karlinsky.
\newblock Teaching structured vision language concepts to vision language models, 2022.

\bibitem[Doveh et~al.(2023)Doveh, Arbelle, Harary, Herzig, Kim, Cascante-bonilla, Alfassy, Panda, Giryes, Feris, Ullman, and Karlinsky]{DAC}
Sivan Doveh, Assaf Arbelle, Sivan Harary, Roei Herzig, Donghyun Kim, Paola Cascante-bonilla, Amit Alfassy, Rameswar Panda, Raja Giryes, Rogerio Feris, Shimon Ullman, and Leonid Karlinsky.
\newblock Dense and aligned captions (dac) promote compositional reasoning in vl models, 2023.

\bibitem[Gadre et~al.(2023)Gadre, Ilharco, Fang, Hayase, Smyrnis, Nguyen, Marten, Wortsman, Ghosh, Zhang, Orgad, Entezari, Daras, Pratt, Ramanujan, Bitton, Marathe, Mussmann, Vencu, Cherti, Krishna, Koh, Saukh, Ratner, Song, Hajishirzi, Farhadi, Beaumont, Oh, Dimakis, Jitsev, Carmon, Shankar, and Schmidt]{gadre2023datacomp}
Samir~Yitzhak Gadre, Gabriel Ilharco, Alex Fang, Jonathan Hayase, Georgios Smyrnis, Thao Nguyen, Ryan Marten, Mitchell Wortsman, Dhruba Ghosh, Jieyu Zhang, Eyal Orgad, Rahim Entezari, Giannis Daras, Sarah Pratt, Vivek Ramanujan, Yonatan Bitton, Kalyani Marathe, Stephen Mussmann, Richard Vencu, Mehdi Cherti, Ranjay Krishna, Pang~Wei Koh, Olga Saukh, Alexander Ratner, Shuran Song, Hannaneh Hajishirzi, Ali Farhadi, Romain Beaumont, Sewoong Oh, Alex Dimakis, Jenia Jitsev, Yair Carmon, Vaishaal Shankar, and Ludwig Schmidt.
\newblock Datacomp: In search of the next generation of multimodal datasets, 2023.

\bibitem[Gao et~al.(2023)Gao, Geng, Zhang, Ma, Fang, Zhang, Li, and Qiao]{clip-adapter}
Peng Gao, Shijie Geng, Renrui Zhang, Teli Ma, Rongyao Fang, Yongfeng Zhang, Hongsheng Li, and Yu Qiao.
\newblock Clip-adapter: Better vision-language models with feature adapters.
\newblock \emph{International Journal of Computer Vision}, pages 1--15, 2023.

\bibitem[Ge et~al.(2018)Ge, Huang, Dong, and Scott]{Ge2018DeepML}
Weifeng Ge, Weilin Huang, Dengke Dong, and Matthew~R. Scott.
\newblock Deep metric learning with hierarchical triplet loss.
\newblock In \emph{European Conference on Computer Vision}, 2018.

\bibitem[Goel et~al.(2022)Goel, Bansal, Bhatia, Rossi, Vinay, and Grover]{goel2022cyclip}
Shashank Goel, Hritik Bansal, Sumit Bhatia, Ryan~A. Rossi, Vishwa Vinay, and Aditya Grover.
\newblock Cyclip: Cyclic contrastive language-image pretraining, 2022.

\bibitem[Goyal et~al.(2017{\natexlab{a}})Goyal, Khot, Summers-Stay, Batra, and Parikh]{vqa2017}
Yash Goyal, Tejas Khot, Douglas Summers-Stay, Dhruv Batra, and Devi Parikh.
\newblock Making the v in vqa matter: Elevating the role of image understanding in visual question answering, 2017{\natexlab{a}}.

\bibitem[Goyal et~al.(2017{\natexlab{b}})Goyal, Khot, Summers-Stay, Batra, and Parikh]{vqav2}
Yash Goyal, Tejas Khot, Douglas Summers-Stay, Dhruv Batra, and Devi Parikh.
\newblock Making the v in vqa matter: Elevating the role of image understanding in visual question answering, 2017{\natexlab{b}}.

\bibitem[Harwood et~al.(2017)Harwood, Kumar, Carneiro, Reid, and Drummond]{Harwood2017SmartMF}
Ben Harwood, B.~V. Kumar, G. Carneiro, Ian~D. Reid, and Tom Drummond.
\newblock Smart mining for deep metric learning.
\newblock \emph{2017 IEEE International Conference on Computer Vision (ICCV)}, pages 2840--2848, 2017.

\bibitem[Herzig et~al.(2023)Herzig, Mendelson, Karlinsky, Arbelle, Feris, Darrell, and Globerson]{SGVL}
Roei Herzig, Alon Mendelson, Leonid Karlinsky, Assaf Arbelle, Rogerio Feris, Trevor Darrell, and Amir Globerson.
\newblock Incorporating structured representations into pretrained vision - language models using scene graphs, 2023.

\bibitem[Hessel et~al.(2022)Hessel, Holtzman, Forbes, Bras, and Choi]{hessel2022clipscore}
Jack Hessel, Ari Holtzman, Maxwell Forbes, Ronan~Le Bras, and Yejin Choi.
\newblock Clipscore: A reference-free evaluation metric for image captioning, 2022.

\bibitem[Honnibal and Montani(2017)]{spacy2}
Matthew Honnibal and Ines Montani.
\newblock {spaCy 2}: Natural language understanding with {B}loom embeddings, convolutional neural networks and incremental parsing.
\newblock To appear, 2017.

\bibitem[Hsieh et~al.(2023)Hsieh, Zhang, Ma, Kembhavi, and Krishna]{hsieh2023sugarcrepe}
Cheng-Yu Hsieh, Jieyu Zhang, Zixian Ma, Aniruddha Kembhavi, and Ranjay Krishna.
\newblock Sugarcrepe: Fixing hackable benchmarks for vision-language compositionality, 2023.

\bibitem[Jia et~al.(2021)Jia, Yang, Xia, Chen, Parekh, Pham, Le, Sung, Li, and Duerig]{align}
Chao Jia, Yinfei Yang, Ye Xia, Yi-Ting Chen, Zarana Parekh, Hieu Pham, Quoc~V. Le, Yunhsuan Sung, Zhen Li, and Tom Duerig.
\newblock Scaling up visual and vision-language representation learning with noisy text supervision, 2021.

\bibitem[Kalantidis et~al.(2020{\natexlab{a}})Kalantidis, Sariyildiz, Pion, Weinzaepfel, and Larlus]{Kalantidis2020HardNM}
Yannis Kalantidis, Mert~Bulent Sariyildiz, No'e Pion, Philippe Weinzaepfel, and Diane Larlus.
\newblock Hard negative mixing for contrastive learning.
\newblock \emph{ArXiv}, abs/2010.01028, 2020{\natexlab{a}}.

\bibitem[Kalantidis et~al.(2020{\natexlab{b}})Kalantidis, Sariyildiz, Pion, Weinzaepfel, and Larlus]{kalantidis2020hard}
Yannis Kalantidis, Mert~Bulent Sariyildiz, Noe Pion, Philippe Weinzaepfel, and Diane Larlus.
\newblock Hard negative mixing for contrastive learning.
\newblock \emph{Advances in Neural Information Processing Systems}, 33:\penalty0 21798--21809, 2020{\natexlab{b}}.

\bibitem[Kirillov et~al.(2023)Kirillov, Mintun, Ravi, Mao, Rolland, Gustafson, Xiao, Whitehead, Berg, Lo, et~al.]{sam}
Alexander Kirillov, Eric Mintun, Nikhila Ravi, Hanzi Mao, Chloe Rolland, Laura Gustafson, Tete Xiao, Spencer Whitehead, Alexander~C Berg, Wan-Yen Lo, et~al.
\newblock Segment anything.
\newblock \emph{arXiv preprint arXiv:2304.02643}, 2023.

\bibitem[Krishna et~al.(2016)Krishna, Zhu, Groth, Johnson, Hata, Kravitz, Chen, Kalantidis, Li, Shamma, Bernstein, and Li]{visualgenome}
Ranjay Krishna, Yuke Zhu, Oliver Groth, Justin Johnson, Kenji Hata, Joshua Kravitz, Stephanie Chen, Yannis Kalantidis, Li-Jia Li, David~A. Shamma, Michael~S. Bernstein, and Fei-Fei Li.
\newblock Visual genome: Connecting language and vision using crowdsourced dense image annotations, 2016.

\bibitem[Krizhevsky et~al.()Krizhevsky, Nair, and Hinton]{cifar}
Alex Krizhevsky, Vinod Nair, and Geoffrey Hinton.
\newblock Cifar-10 (canadian institute for advanced research).

\bibitem[Li et~al.(2021)Li, Selvaraju, Gotmare, Joty, Xiong, and Hoi]{albef}
Junnan Li, Ramprasaath~R. Selvaraju, Akhilesh~Deepak Gotmare, Shafiq Joty, Caiming Xiong, and Steven Hoi.
\newblock Align before fuse: Vision and language representation learning with momentum distillation, 2021.

\bibitem[Li et~al.(2022)Li, Li, Xiong, and Hoi]{blip}
Junnan Li, Dongxu Li, Caiming Xiong, and Steven Hoi.
\newblock Blip: Bootstrapping language-image pre-training for unified vision-language understanding and generation, 2022.

\bibitem[Li et~al.(2023)Li, Li, Savarese, and Hoi]{blip2}
Junnan Li, Dongxu Li, Silvio Savarese, and Steven Hoi.
\newblock Blip-2: Bootstrapping language-image pre-training with frozen image encoders and large language models, 2023.

\bibitem[Lin et~al.(2014)Lin, Maire, Belongie, Hays, Perona, Ramanan, Doll{\'a}r, and Zitnick]{mscoco}
Tsung-Yi Lin, Michael Maire, Serge Belongie, James Hays, Pietro Perona, Deva Ramanan, Piotr Doll{\'a}r, and C~Lawrence Zitnick.
\newblock Microsoft coco: Common objects in context.
\newblock In \emph{Computer Vision--ECCV 2014: 13th European Conference, Zurich, Switzerland, September 6-12, 2014, Proceedings, Part V 13}, pages 740--755. Springer, 2014.

\bibitem[Lin et~al.(2015)Lin, Maire, Belongie, Bourdev, Girshick, Hays, Perona, Ramanan, Zitnick, and Dollár]{coco2015}
Tsung-Yi Lin, Michael Maire, Serge Belongie, Lubomir Bourdev, Ross Girshick, James Hays, Pietro Perona, Deva Ramanan, C.~Lawrence Zitnick, and Piotr Dollár.
\newblock Microsoft coco: Common objects in context, 2015.

\bibitem[Liu et~al.(2019{\natexlab{a}})Liu, Zhu, Lei, and Li]{Liu_2019_CVPR}
Hao Liu, Xiangyu Zhu, Zhen Lei, and Stan~Z. Li.
\newblock Adaptiveface: Adaptive margin and sampling for face recognition.
\newblock In \emph{Proceedings of the IEEE/CVF Conference on Computer Vision and Pattern Recognition (CVPR)}, 2019{\natexlab{a}}.

\bibitem[Liu et~al.(2023{\natexlab{a}})Liu, Li, Wu, and Lee]{llava}
Haotian Liu, Chunyuan Li, Qingyang Wu, and Yong~Jae Lee.
\newblock Visual instruction tuning, 2023{\natexlab{a}}.

\bibitem[Liu et~al.(2023{\natexlab{b}})Liu, Wang, Wang, Smith, Choi, and Hajishirzi]{liu2023vera}
Jiacheng Liu, Wenya Wang, Dianzhuo Wang, Noah~A. Smith, Yejin Choi, and Hannaneh Hajishirzi.
\newblock Vera: A general-purpose plausibility estimation model for commonsense statements, 2023{\natexlab{b}}.

\bibitem[Liu et~al.(2019{\natexlab{b}})Liu, Ott, Goyal, Du, Joshi, Chen, Levy, Lewis, Zettlemoyer, and Stoyanov]{liu2019roberta}
Yinhan Liu, Myle Ott, Naman Goyal, Jingfei Du, Mandar Joshi, Danqi Chen, Omer Levy, Mike Lewis, Luke Zettlemoyer, and Veselin Stoyanov.
\newblock Roberta: A robustly optimized bert pretraining approach, 2019{\natexlab{b}}.

\bibitem[Ma et~al.(2022)Ma, Hong, Gul, Gandhi, Gao, and Krishna]{ma2022crepe}
Zixian Ma, Jerry Hong, Mustafa~Omer Gul, Mona Gandhi, Irena Gao, and Ranjay Krishna.
\newblock Crepe: Can vision-language foundation models reason compositionally?
\newblock \emph{arXiv preprint arXiv:2212.07796}, 2022.

\bibitem[Manmatha et~al.(2017)Manmatha, Wu, Smola, and Kr{\"a}henb{\"u}hl]{Manmatha2017SamplingMI}
R. Manmatha, Chaoxia Wu, Alex Smola, and Philipp Kr{\"a}henb{\"u}hl.
\newblock Sampling matters in deep embedding learning.
\newblock \emph{2017 IEEE International Conference on Computer Vision (ICCV)}, pages 2859--2867, 2017.

\bibitem[Mañas et~al.(2023{\natexlab{a}})Mañas, Rodriguez, Ahmadi, Nematzadeh, Goyal, and Agrawal]{mapl}
Oscar Mañas, Pau Rodriguez, Saba Ahmadi, Aida Nematzadeh, Yash Goyal, and Aishwarya Agrawal.
\newblock Mapl: Parameter-efficient adaptation of unimodal pre-trained models for vision-language few-shot prompting, 2023{\natexlab{a}}.

\bibitem[Mañas et~al.(2023{\natexlab{b}})Mañas, Rodriguez, Ahmadi, Nematzadeh, Goyal, and Agrawal]{mañas2023mapl}
Oscar Mañas, Pau Rodriguez, Saba Ahmadi, Aida Nematzadeh, Yash Goyal, and Aishwarya Agrawal.
\newblock Mapl: Parameter-efficient adaptation of unimodal pre-trained models for vision-language few-shot prompting, 2023{\natexlab{b}}.

\bibitem[Metzen et~al.(2023)Metzen, Saranrittichai, and Mummadi]{Metzen2023AutoCLIPAZ}
Jan~Hendrik Metzen, Piyapat Saranrittichai, and Chaithanya~Kumar Mummadi.
\newblock Autoclip: Auto-tuning zero-shot classifiers for vision-language models.
\newblock \emph{ArXiv}, abs/2309.16414, 2023.

\bibitem[Minderer et~al.(2022)Minderer, Gritsenko, Stone, Neumann, Weissenborn, Dosovitskiy, Mahendran, Arnab, Dehghani, Shen, Wang, Zhai, Kipf, and Houlsby]{OWL-ViT}
Matthias Minderer, Alexey Gritsenko, Austin Stone, Maxim Neumann, Dirk Weissenborn, Alexey Dosovitskiy, Aravindh Mahendran, Anurag Arnab, Mostafa Dehghani, Zhuoran Shen, Xiao Wang, Xiaohua Zhai, Thomas Kipf, and Neil Houlsby.
\newblock Simple open-vocabulary object detection with vision transformers, 2022.

\bibitem[Morris et~al.(2020)Morris, Lifland, Yoo, Grigsby, Jin, and Qi]{grammar}
John~X. Morris, Eli Lifland, Jin~Yong Yoo, Jake Grigsby, Di Jin, and Yanjun Qi.
\newblock Textattack: A framework for adversarial attacks, data augmentation, and adversarial training in nlp, 2020.

\bibitem[Novack et~al.(2023)Novack, Garg, McAuley, and Lipton]{Novack2023CHiLSZI}
Zachary Novack, S. Garg, Julian McAuley, and Zachary~Chase Lipton.
\newblock Chils: Zero-shot image classification with hierarchical label sets.
\newblock \emph{ArXiv}, abs/2302.02551, 2023.

\bibitem[Paiss et~al.(2023)Paiss, Ephrat, Tov, Zada, Mosseri, Irani, and Dekel]{paiss2023teaching}
Roni Paiss, Ariel Ephrat, Omer Tov, Shiran Zada, Inbar Mosseri, Michal Irani, and Tali Dekel.
\newblock Teaching clip to count to ten, 2023.

\bibitem[Parcalabescu et~al.(2022)Parcalabescu, Cafagna, Muradjan, Frank, Calixto, and Gatt]{parcalabescu-etal-2022-valse}
Letitia Parcalabescu, Michele Cafagna, Lilitta Muradjan, Anette Frank, Iacer Calixto, and Albert Gatt.
\newblock {VALSE}: A task-independent benchmark for vision and language models centered on linguistic phenomena.
\newblock In \emph{Proceedings of the 60th Annual Meeting of the Association for Computational Linguistics (Volume 1: Long Papers)}, pages 8253--8280, Dublin, Ireland, 2022. Association for Computational Linguistics.

\bibitem[Poole et~al.(2022)Poole, Jain, Barron, and Mildenhall]{poole2022dreamfusion}
Ben Poole, Ajay Jain, Jonathan~T Barron, and Ben Mildenhall.
\newblock Dreamfusion: Text-to-3d using 2d diffusion.
\newblock \emph{arXiv preprint arXiv:2209.14988}, 2022.

\bibitem[Qin et~al.(2021)Qin, Zhang, Chen, Lakshminarayanan, Beutel, and Wang]{Qin2021UnderstandingAI}
Yao Qin, Chiyuan Zhang, Ting Chen, Balaji Lakshminarayanan, Alex Beutel, and Xuezhi Wang.
\newblock Understanding and improving robustness of vision transformers through patch-based negative augmentation.
\newblock \emph{ArXiv}, abs/2110.07858, 2021.

\bibitem[Radford et~al.(2021)Radford, Kim, Hallacy, Ramesh, Goh, Agarwal, Sastry, Askell, Mishkin, Clark, et~al.]{radford2021learning}
Alec Radford, Jong~Wook Kim, Chris Hallacy, Aditya Ramesh, Gabriel Goh, Sandhini Agarwal, Girish Sastry, Amanda Askell, Pamela Mishkin, Jack Clark, et~al.
\newblock Learning transferable visual models from natural language supervision.
\newblock In \emph{International conference on machine learning}, pages 8748--8763. PMLR, 2021.

\bibitem[Ramesh et~al.(2022)Ramesh, Dhariwal, Nichol, Chu, and Chen]{dalle}
Aditya Ramesh, Prafulla Dhariwal, Alex Nichol, Casey Chu, and Mark Chen.
\newblock Hierarchical text-conditional image generation with clip latents, 2022.

\bibitem[Robinson et~al.(2020)Robinson, Chuang, Sra, and Jegelka]{Robinson2020ContrastiveLW}
Joshua Robinson, Ching-Yao Chuang, Suvrit Sra, and Stefanie Jegelka.
\newblock Contrastive learning with hard negative samples.
\newblock \emph{ArXiv}, abs/2010.04592, 2020.

\bibitem[Rombach et~al.(2022)Rombach, Blattmann, Lorenz, Esser, and Ommer]{latentdiffusion}
Robin Rombach, Andreas Blattmann, Dominik Lorenz, Patrick Esser, and Bj{\"o}rn Ommer.
\newblock High-resolution image synthesis with latent diffusion models.
\newblock In \emph{Proceedings of the IEEE/CVF conference on computer vision and pattern recognition}, pages 10684--10695, 2022.

\bibitem[Schuhmann et~al.(2021)Schuhmann, Vencu, Beaumont, Kaczmarczyk, Mullis, Katta, Coombes, Jitsev, and Komatsuzaki]{schuhmann2021laion400m}
Christoph Schuhmann, Richard Vencu, Romain Beaumont, Robert Kaczmarczyk, Clayton Mullis, Aarush Katta, Theo Coombes, Jenia Jitsev, and Aran Komatsuzaki.
\newblock Laion-400m: Open dataset of clip-filtered 400 million image-text pairs, 2021.

\bibitem[Schuhmann et~al.(2022)Schuhmann, Beaumont, Vencu, Gordon, Wightman, Cherti, Coombes, Katta, Mullis, Wortsman, Schramowski, Kundurthy, Crowson, Schmidt, Kaczmarczyk, and Jitsev]{schuhmann2022laion5b}
Christoph Schuhmann, Romain Beaumont, Richard Vencu, Cade Gordon, Ross Wightman, Mehdi Cherti, Theo Coombes, Aarush Katta, Clayton Mullis, Mitchell Wortsman, Patrick Schramowski, Srivatsa Kundurthy, Katherine Crowson, Ludwig Schmidt, Robert Kaczmarczyk, and Jenia Jitsev.
\newblock Laion-5b: An open large-scale dataset for training next generation image-text models, 2022.

\bibitem[Sharma et~al.(2018)Sharma, Ding, Goodman, and Soricut]{Sharma2018ConceptualCA}
Piyush Sharma, Nan Ding, Sebastian Goodman, and Radu Soricut.
\newblock Conceptual captions: A cleaned, hypernymed, image alt-text dataset for automatic image captioning.
\newblock In \emph{Annual Meeting of the Association for Computational Linguistics}, 2018.

\bibitem[Singh et~al.(2022)Singh, Hu, Goswami, Couairon, Galuba, Rohrbach, and Kiela]{singh2022flava}
Amanpreet Singh, Ronghang Hu, Vedanuj Goswami, Guillaume Couairon, Wojciech Galuba, Marcus Rohrbach, and Douwe Kiela.
\newblock Flava: A foundational language and vision alignment model, 2022.

\bibitem[Singh et~al.(2023)Singh, Zhang, Wang, Wang, Xiong, Du, and Chen]{Singh2023CoarsetoFineCL}
Harman Singh, Pengchuan Zhang, Qifan Wang, Mengjiao Wang, Wenhan Xiong, Jingfei Du, and Yu Chen.
\newblock Coarse-to-fine contrastive learning in image-text-graph space for improved vision-language compositionality.
\newblock 2023.

\bibitem[Sun et~al.(2023)Sun, Fang, Wu, Wang, and Cao]{sun2023evaclip}
Quan Sun, Yuxin Fang, Ledell Wu, Xinlong Wang, and Yue Cao.
\newblock Eva-clip: Improved training techniques for clip at scale, 2023.

\bibitem[Tan and Bansal(2019)]{tan2019lxmert}
Hao Tan and Mohit Bansal.
\newblock Lxmert: Learning cross-modality encoder representations from transformers, 2019.

\bibitem[Thrush et~al.(2022)Thrush, Jiang, Bartolo, Singh, Williams, Kiela, and Ross]{thrush2022winoground}
Tristan Thrush, Ryan Jiang, Max Bartolo, Amanpreet Singh, Adina Williams, Douwe Kiela, and Candace Ross.
\newblock Winoground: Probing vision and language models for visio-linguistic compositionality.
\newblock In \emph{Proceedings of the IEEE/CVF Conference on Computer Vision and Pattern Recognition}, pages 5238--5248, 2022.

\bibitem[Wang et~al.(2023)Wang, Vasu, Faghri, Vemulapalli, Farajtabar, Mehta, Rastegari, Tuzel, and Pouransari]{wang2023samclip}
Haoxiang Wang, Pavan Kumar~Anasosalu Vasu, Fartash Faghri, Raviteja Vemulapalli, Mehrdad Farajtabar, Sachin Mehta, Mohammad Rastegari, Oncel Tuzel, and Hadi Pouransari.
\newblock Sam-clip: Merging vision foundation models towards semantic and spatial understanding, 2023.

\bibitem[Wang et~al.(2014)Wang, song, Leung, Rosenberg, Wang, Philbin, Chen, and Wu]{wang2014learning}
Jiang Wang, Yang song, Thomas Leung, Chuck Rosenberg, Jinbin Wang, James Philbin, Bo Chen, and Ying Wu.
\newblock Learning fine-grained image similarity with deep ranking, 2014.

\bibitem[Wang et~al.(2022)Wang, Bao, Dong, Bjorck, Peng, Liu, Aggarwal, Mohammed, Singhal, Som, and Wei]{beit3}
Wenhui Wang, Hangbo Bao, Li Dong, Johan Bjorck, Zhiliang Peng, Qiang Liu, Kriti Aggarwal, Owais~Khan Mohammed, Saksham Singhal, Subhojit Som, and Furu Wei.
\newblock Image as a foreign language: Beit pretraining for all vision and vision-language tasks, 2022.

\bibitem[Wang et~al.(2021)Wang, Chen, and Zhu]{wang2021survey}
Xin Wang, Yudong Chen, and Wenwu Zhu.
\newblock A survey on curriculum learning, 2021.

\bibitem[Xu et~al.(2021)Xu, Zhang, Wei, Lin, Cao, Hu, and Bai]{Xu2021ASB}
Mengde Xu, Zheng Zhang, Fangyun Wei, Yutong Lin, Yue Cao, Han Hu, and Xiang Bai.
\newblock A simple baseline for zero-shot semantic segmentation with pre-trained vision-language model.
\newblock \emph{ArXiv}, abs/2112.14757, 2021.

\bibitem[Yu et~al.(2022)Yu, Wang, Vasudevan, Yeung, Seyedhosseini, and Wu]{coca}
Jiahui Yu, Zirui Wang, Vijay Vasudevan, Legg Yeung, Mojtaba Seyedhosseini, and Yonghui Wu.
\newblock Coca: Contrastive captioners are image-text foundation models, 2022.

\bibitem[Yuksekgonul et~al.(2022)Yuksekgonul, Bianchi, Kalluri, Jurafsky, and Zou]{yuksekgonul2022and}
Mert Yuksekgonul, Federico Bianchi, Pratyusha Kalluri, Dan Jurafsky, and James Zou.
\newblock When and why vision-language models behave like bags-of-words, and what to do about it?
\newblock \emph{arXiv e-prints}, pages arXiv--2210, 2022.

\bibitem[Zeng et~al.(2022)Zeng, Zhang, and Li]{zeng2022multigrained}
Yan Zeng, Xinsong Zhang, and Hang Li.
\newblock Multi-grained vision language pre-training: Aligning texts with visual concepts, 2022.

\bibitem[Zhai et~al.(2023)Zhai, Mustafa, Kolesnikov, and Beyer]{zhai2023sigmoid}
Xiaohua Zhai, Basil Mustafa, Alexander Kolesnikov, and Lucas Beyer.
\newblock Sigmoid loss for language image pre-training, 2023.

\bibitem[Zhang et~al.(2018)Zhang, Ouyang, Li, and Xu]{zhang2018collaborative}
Weichen Zhang, Wanli Ouyang, Wen Li, and Dong Xu.
\newblock Collaborative and adversarial network for unsupervised domain adaptation.
\newblock In \emph{Proceedings of the IEEE conference on computer vision and pattern recognition}, pages 3801--3809, 2018.

\bibitem[Zhao et~al.(2022)Zhao, Zhang, Zhu, Shen, Lee, Lu, and Yin]{zhao2022vlchecklist}
Tiancheng Zhao, Tianqi Zhang, Mingwei Zhu, Haozhan Shen, Kyusong Lee, Xiaopeng Lu, and Jianwei Yin.
\newblock Vl-checklist: Evaluating pre-trained vision-language models with objects, attributes and relations, 2022.

\bibitem[Zhao et~al.(2019)Zhao, Qi, Luo, and Davis]{zhao2019weakly}
Xiaonan Zhao, Huan Qi, Rui Luo, and Larry Davis.
\newblock A weakly supervised adaptive triplet loss for deep metric learning, 2019.

\bibitem[Zhou et~al.(2022{\natexlab{a}})Zhou, Loy, and Dai]{maskclip}
Chong Zhou, Chen~Change Loy, and Bo Dai.
\newblock Extract free dense labels from clip, 2022{\natexlab{a}}.

\bibitem[Zhou et~al.(2022{\natexlab{b}})Zhou, Yang, Loy, and Liu]{coop}
Kaiyang Zhou, Jingkang Yang, Chen~Change Loy, and Ziwei Liu.
\newblock Learning to prompt for vision-language models.
\newblock \emph{International Journal of Computer Vision}, 130\penalty0 (9):\penalty0 2337--2348, 2022{\natexlab{b}}.

\bibitem[Zhou et~al.(2023)Zhou, Zhang, Lei, Liu, and Liu]{zhou2023zegclip}
Ziqin Zhou, Bowen Zhang, Yinjie Lei, Lingqiao Liu, and Yifan Liu.
\newblock Zegclip: Towards adapting clip for zero-shot semantic segmentation, 2023.

\bibitem[Zhu et~al.(2023{\natexlab{a}})Zhu, Chen, Shen, Li, and Elhoseiny]{zhu2023minigpt}
Deyao Zhu, Jun Chen, Xiaoqian Shen, Xiang Li, and Mohamed Elhoseiny.
\newblock Minigpt-4: Enhancing vision-language understanding with advanced large language models.
\newblock \emph{arXiv preprint arXiv:2304.10592}, 2023{\natexlab{a}}.

\bibitem[Zhu et~al.(2023{\natexlab{b}})Zhu, Chen, Shen, Li, and Elhoseiny]{zhu2023minigpt4}
Deyao Zhu, Jun Chen, Xiaoqian Shen, Xiang Li, and Mohamed Elhoseiny.
\newblock Minigpt-4: Enhancing vision-language understanding with advanced large language models, 2023{\natexlab{b}}.

\end{thebibliography}
}
\clearpage
\setcounter{page}{1}
\maketitlesupplementary

\section{Analysis of learned representations}

\begin{figure}[htb]
    \centering
    
    \label{fig:derivation}
    \includegraphics[width=0.9\linewidth]{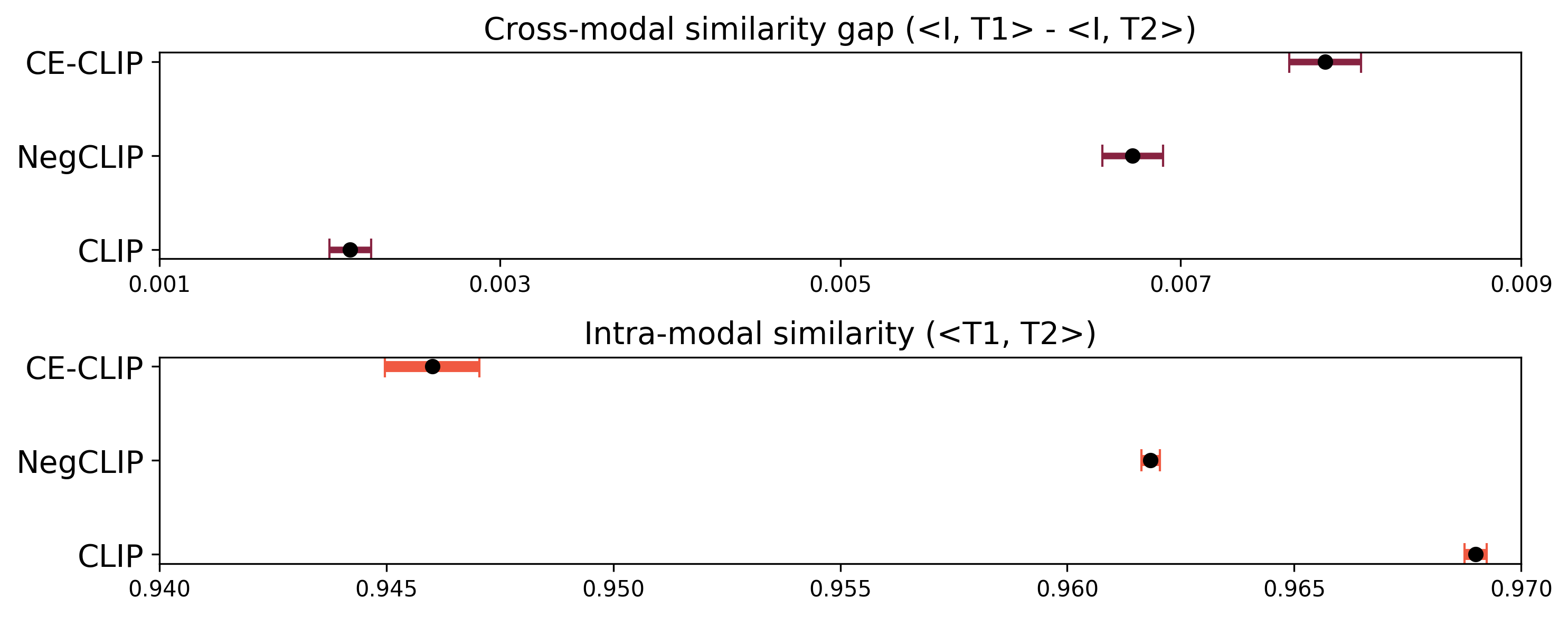}
       \captionsetup{font=small}
     % \caption{\textbf{Bootstrap Sampling for 99\% Confidence Intervals in Standard Deviation Estimation}. T1 is correct caption while T2 is false caption.}
    \caption{Analyzing the intra-modal similarities and cross-modal similarity gaps yielded by different methods on the ARO benchmark. T1 refers to positive caption while T2 refers to hard-negative caption. The red lines denote the standard errors obtained with bootstrapping 50,000 samples with confidence interval of 99\%.}
     \label{fig:intervals}
    \vspace{-5mm}
\end{figure}

In this section, we examine the image and text representations learned by our model. In particular, we investigate whether our method learns more distinct representations for positive and hard negative examples compared to those learned by CLIP and NegCLP. For each of CLIP, NegCLIP and CE-CLIP, we measure the intra-modal similarity between positive and hard-negative captions, as well as, the cross-modal similarity gap between positive and hard-negative image-caption pairs. We expect our method to reduce the intra-modal similarity and enlarge the cross-modal similarity gap compared to CLIP and NegCLIP. We report the results in \cref{fig:intervals}, which shows that CE-CLIP achieves statistically significantly better intra-modal similarity (lower is better) and cross-modal similarity gap (higher is better) compared to CLIP and NegCLIP. To compute statistical significance, we used bootstrapping with 50,000 samples with confidence interval of 99\%. 

\section{Qualitative Examples}

\begin{figure*}[]
    \centering
    \includegraphics[width=0.95\textwidth]{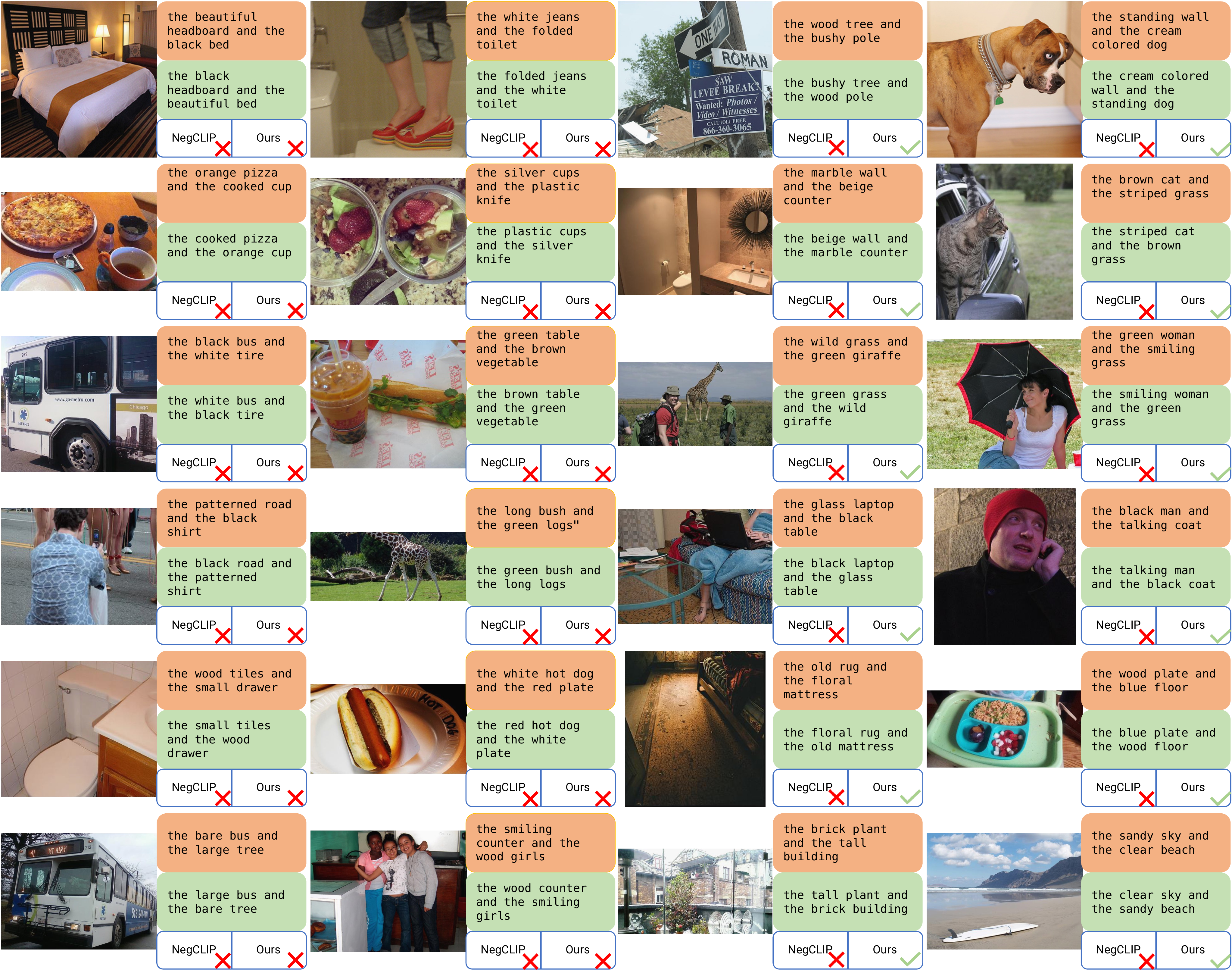}
    \caption{Some qualitative examples from ARO-Attribute. Caption in red box is unmatched and in green box is matched. {\color{green}\ding{51}} represents model predicates correctly and {\color{red}\ding{55}} means wrong.}
    \vspace{-3mm}
    \label{fig:quantiative_attribute}
\end{figure*}

\begin{figure*}[]
    \centering
    \includegraphics[width=0.95\textwidth]{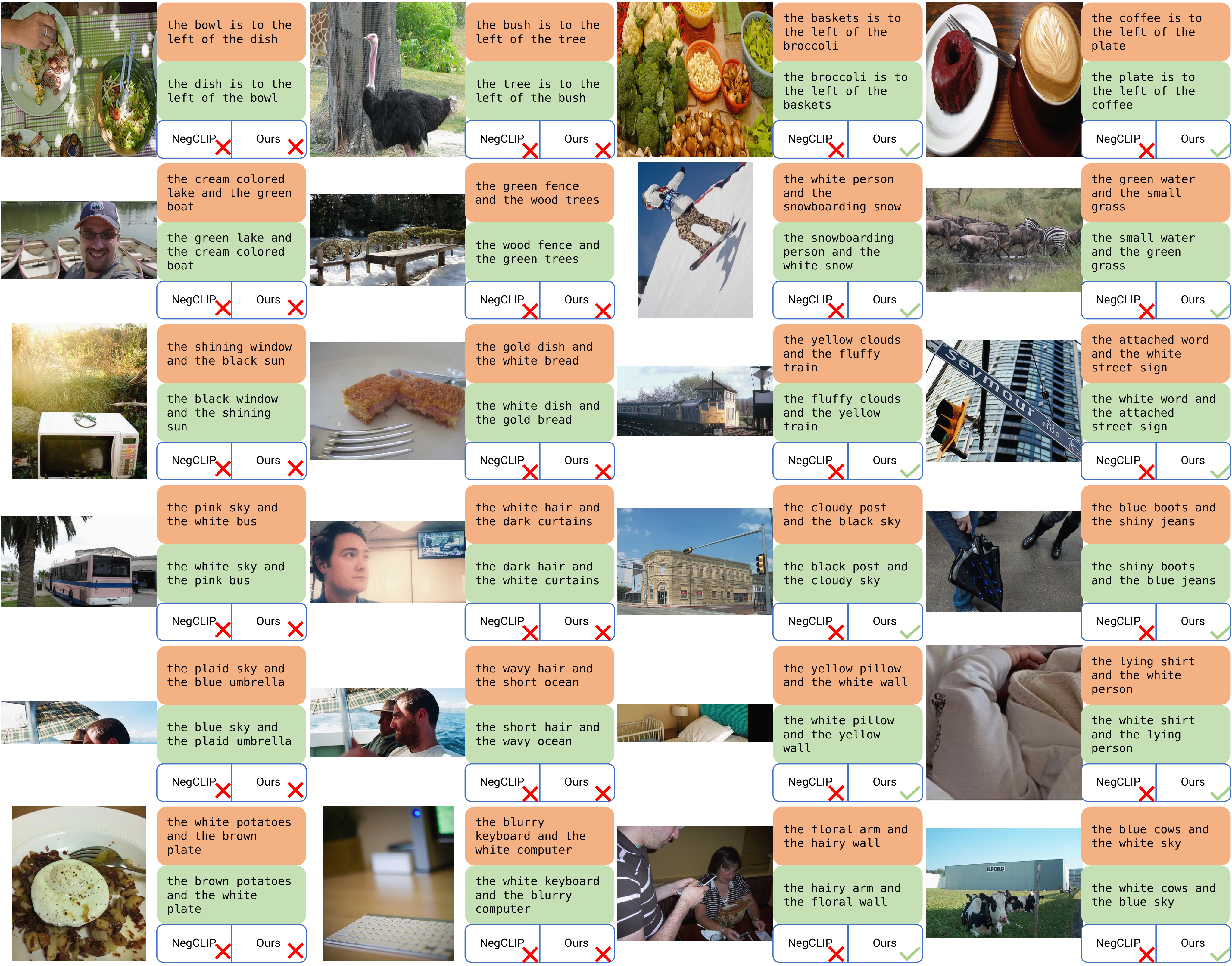}
    \caption{Some qualitative examples from ARO-Relation. Caption in red box is unmatched and in green box is matched. {\color{green}\ding{51}} represents model predicates correctly and {\color{red}\ding{55}} means wrong.}
    \vspace{-3mm}
    \label{fig:quantiative_relation}
\end{figure*}

\cref{fig:quantiative_attribute} and \cref{fig:quantiative_relation} illustrate some side-by-side comparisons of hits and misses by CE-CLIP versus NegCLIP.

\section{Benchmark}
The statistics of benchmarks we use are shown in \cref{tab:statistics}.
\begin{table}[h]
\centering
\resizebox{\linewidth}{!}{
    \begin{tabular}{l c r }
    \toprule
    \textbf{Benchmark} & \textbf{Task} & \textbf{\# image-text pairs}  \\
    \midrule

    ARO-Relation  & Relation             & 24k  \\
    ARO-Attribution & Attribution & 28.7k \\
    VALSE       & Linguistic Phenomena     & 6.8k   \\
    VL-CheckList& Objects, Attributes and Relations & 410k\\
    SugarCrepe & Objects, Attributes and Relations &7.5k\\
    \bottomrule
    \end{tabular}
}

\captionsetup{font=small}
\caption{Overview of vl-compositional benchmarks.}
\label{tab:statistics}
\vspace{-5mm}
\end{table}

\section{Acknowledgement}
We thank the Mila IDT team and their technical support for maintaining the Mila compute cluster. We also acknowledge the material support of NVIDIA in the form of computational resources. Throughout this project, Aishwarya Agrawal received support from the Canada CIFAR AI Chair award.

\end{document}